\documentclass{article} 
\usepackage{iclr2024_conference,times}

\usepackage{hyperref}       
\usepackage{url}            
\usepackage{booktabs}       
\usepackage{multirow}
\usepackage{graphicx}
\usepackage{subcaption}
\usepackage{algorithm}
\usepackage{algpseudocode}

\usepackage{amsmath}
\usepackage{amssymb}
\usepackage{amsthm}
\usepackage{amsfonts}
\usepackage{nicefrac}       
\usepackage{microtype}      
\usepackage{xcolor}         
\usepackage[capitalise]{cleveref}

\definecolor{mydarkblue}{rgb}{0,0.08,0.45}
\hypersetup{colorlinks,linkcolor=red,citecolor=teal,urlcolor=blue}

\def\R{\mathbb{R}}

\def\U{\mathbf{U}}

\def\E{\mathbb{E}}

\def\D{\mathcal{D}}

\def\v{\mathbf{v}}
\def\bS{\boldsymbol{\Sigma}}
\def\bL{\boldsymbol{\Lambda}}
\def\s{\mathbf{s}}

\def\x{\mathbf{x}}
\def\u{\mathbf{u}}

\def\KL{\mathrm{KL}}

\DeclareMathOperator*{\argmin}{argmin}

\title{Input-gradient space particle inference for neural network ensembles}

\author{
  Trung Trinh$^1$ 
  \qquad
  Markus Heinonen$^1$
  \qquad
  Luigi Acerbi$^2$
  \qquad
  Samuel Kaski$^{1,3}$\\
  $^1$Department of Computer Science, Aalto University, Finland\\
  $^2$Department of Computer Science, University of Helsinki, Finland\\
  $^3$Department of Computer Science, University of Manchester, United Kingdom\\
  \texttt{\{trung.trinh, markus.o.heinonen, samuel.kaski\}@aalto.fi},\\ \texttt{luigi.acerbi@helsinki.fi}
}

\iclrfinalcopy 
\begin{document}

\maketitle

\begin{abstract}
Deep Ensembles (DEs) demonstrate improved accuracy, calibration and robustness to perturbations over single neural networks partly due to their functional diversity. 
Particle-based variational inference (ParVI) methods enhance diversity by formalizing a repulsion term based on a network similarity kernel.
However, weight-space repulsion is inefficient due to over-parameterization, while direct function-space repulsion has been found to produce little improvement over DEs.
To sidestep these difficulties, we propose First-order Repulsive Deep Ensemble (FoRDE), an ensemble learning method based on ParVI, which performs repulsion in the space of first-order input gradients.
As input gradients uniquely characterize a function up to translation and are much smaller in dimension than the weights, this method guarantees that ensemble members are functionally different.
Intuitively, diversifying the input gradients encourages each network to learn different features, which is expected to improve the robustness of an ensemble.
Experiments on image classification datasets and transfer learning tasks show that FoRDE 
significantly outperforms the gold-standard DEs and other ensemble methods in accuracy and calibration under covariate shift due to input perturbations.
\end{abstract}

\section{Introduction}
Ensemble methods, which combine predictions from multiple models, are a well-known strategy in machine learning \citep{emsembleinML} to boost predictive performance \citep{lakshminarayanan2017simple}, uncertainty estimation \citep{ovadia2019can}, robustness to adversarial attacks \citep{pmlr-v97-pang19a} and corruptions \citep{hendrycks2018benchmarking}. 
Deep ensembles (DEs) combine multiple neural networks from independent weight initializations \citep{lakshminarayanan2017simple}.
While DEs are simple to implement and have promising performance, their weight-based diversity does not necessarily translate into useful functional diversity \citep{rame2021dice,d'angelo2021repulsive,yashima22a}.

\begin{figure}[t]
    \centering
    \includegraphics[width=\textwidth]{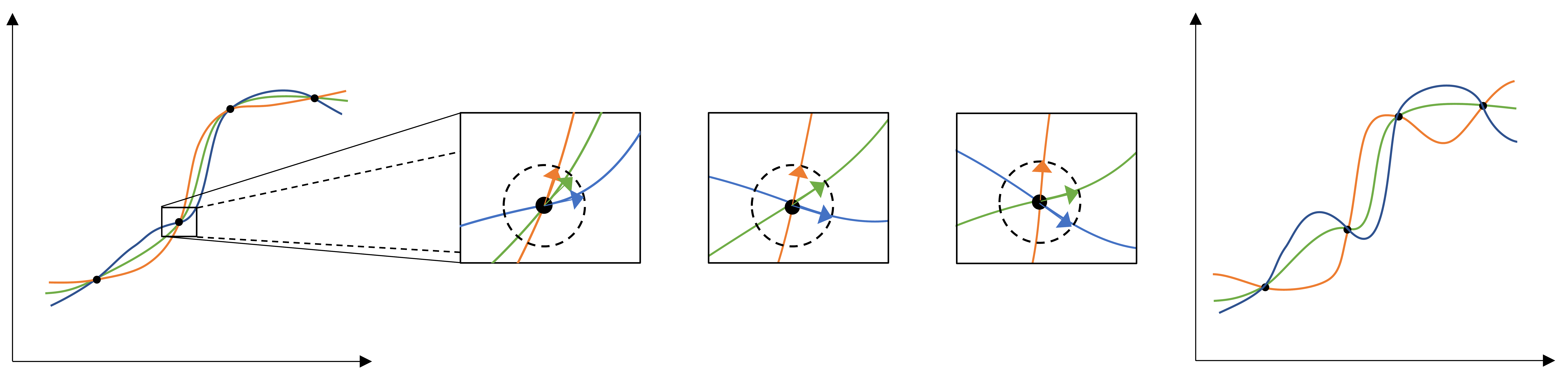}
    \caption{\textbf{Input-gradient repulsion increases functional diversity.} An illustration of input gradient repulsion in 1D regression with 3 neural networks. \textbf{Left:} At some point during training, the models fit well to the training samples yet exhibit low functional diversity. \textbf{Middle}: As training proceeds, at each data point, the repulsion term gradually pushes the input gradients (represented by the arrows) away from each other on a unit sphere. \textbf{Right:} As a result, at the end of training, the ensemble has gained functional diversity.}
    \label{fig:forde_illustration}
\end{figure}

Particle-based variational inference (ParVI) \citep{qiang2016svgd,chen2018unified,liu2019povi,shen2021} has recently emerged as a direction to promote diversity in neural ensembles from the Bayesian perspective \citep{wang2018function,d'angelo2021repulsive}.
Notably, the ParVI update rule adds a kernelized repulsion term $k(f,f')$ between the ensemble networks $f,f'$ for explicit control of the ensemble diversity. 
Typically repulsion is done in the weight space to capture different regions in the weight posterior.
However, due to the over-parameterization of neural networks, weight-space repulsion suffers from redundancy. 
An alternative approach is to define the repulsion in function space \citep{wang2018function,d'angelo2021repulsive}, which requires the challenging computation of a kernel between functions.
Previous works avoided this issue by comparing functions only on training inputs, which leads to underfitting \citep{d2021stein}.
Neither weight nor function space repulsion has led to significant improvements over vanilla DEs.

\begin{figure}[t]
    \centering    
    \includegraphics[width=\textwidth]{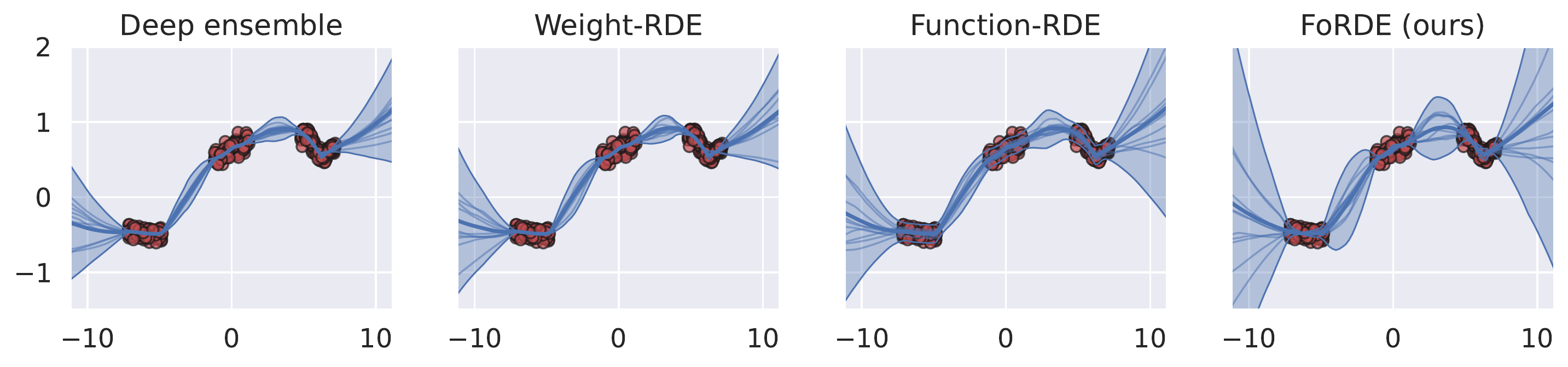}
    \caption{\textbf{Input gradient ensembles (FoRDE) capture higher uncertainty than baselines.} Each panel shows predictive uncertainty in 1D regression for different (repulsive) deep ensemble methods.}
    \label{fig:1D_regression}
\end{figure}

From a functional perspective, a model can \emph{also} be uniquely represented, up to translation, using its first-order derivatives, i.e., input \emph{gradients} $\nabla_\x f$.
Promoting diversity in this third view of input gradients has notable advantages:
\begin{enumerate}
    \item each ensemble member is guaranteed to correspond to a different function;
    \item input gradients have smaller dimensions than weights and thus are more amenable to kernel comparisons;  
    \item unlike function-space repulsion, input-gradient repulsion does not lead to training point underfitting (See \cref{fig:forde_illustration} and the last panel of \cref{fig:1D_regression});
    \item each ensemble member is encouraged to learn different features, which can improve robustness.
\end{enumerate}

In this work, we propose a ParVI neural network ensemble that promotes diversity in their input gradients, called First-order Repulsive deep ensemble (FoRDE). Furthermore, we devise a data-dependent kernel that allows FoRDE to outperform other ensemble methods under input corruptions on image classification tasks. Our code is available at \url{https://github.com/AaltoPML/FoRDE}.

\section{Background}\label{sec:background}

\paragraph{Bayesian neural networks}

In a Bayesian neural network (BNN), we treat the model's weights $\theta$ as random variables with a prior $p(\theta)$. Given a dataset $\D=\{(\x_n,y_n)\}_{n=1}^N$ and a likelihood function $p(y|\x,\theta)$ per data point, we infer the posterior over weights $p(\theta|\D)$ using Bayes' rule
\begin{equation}\label{eq:bnn_posterior}
    p(\theta|\D) = \frac{p(\D|\theta)p(\theta)}{\int_{\theta} p(\D|\theta)p(\theta)\mathrm{d}\theta} = \frac{p(\theta) \prod_{n=1}^N p(y_n|\x_n,\theta)}{\int_{\theta} p(\theta) \prod_{n=1}^N p(y_n|\x_n,\theta) \mathrm{d}\theta},
\end{equation}
where the likelihood is assumed to factorize over data.
To make a prediction on a test sample $\x^*$, we integrate over the inferred posterior in \cref{eq:bnn_posterior}, a practice called \emph{Bayesian model averaging} (BMA):
\begin{equation}\label{eq:bnn_ppd}
    p(y|\x^*, \D) = \int_\theta p(y|\x^*,\theta)p(\theta|\D)\mathrm{d}\theta = \E_{p(\theta|\D)}\big[ p(y|\x^*,\theta) \big].
\end{equation}
However, computing the integral in \cref{eq:bnn_ppd} is intractable for BNNs. Various approximate inference methods have been developed for BNNs, including variational inference (VI) \citep{graves2011practical,blundell2015weight}, Markov chain Monte Carlo (MCMC) \citep{neal2012bayesian,welling2011bayesian,Zhang2020Cyclical} and more recently ParVI \citep{qiang2016svgd,wang2018function,d'angelo2021repulsive}.

\paragraph{Deep ensembles} 

As opposed to BNNs, which attempt to learn the posterior distribution, DEs \citep{lakshminarayanan2017simple} consist of multiple maximum-a-posteriori (MAP) estimates trained from independent random initializations. They can capture diverse functions that explain the data well, as independent training runs under different random conditions will likely converge to different modes in the posterior landscape. DEs have been shown to be better than BNNs in both accuracy and uncertainty estimation \citep{ovadia2019can,Ashukha2020Pitfalls,gustafsson2020evaluating}.

\paragraph{Particle-based variational inference for neural network ensembles} 
ParVI methods \citep{qiang2016svgd,chen2018unified,liu2019povi,shen2021} have been studied recently to formalize neural network ensembles. 
They approximate the target posterior using a set of samples, or particles, by deterministically transporting these particles to the target distribution \citep{qiang2016svgd}.
ParVI methods are expected to be more efficient than MCMC as they take into account the interactions between particles in their update rules \citep{liu2019povi}.
These repulsive interactions are driven by a kernel which measures the pairwise similarities between particles, i.e., networks \citep{liu2019povi}.

The current approaches compare networks in weight space $\theta$ or in function space $f(\cdot ; \theta)$.
%
Weight-space repulsion is ineffective due to difficulties in comparing extremely high-dimensional weight vectors and the existence of weight symmetries  \citep{fort2019deep,entezari2022the}.
Previous studies show that weight-space ParVI does not improve performance over plain DEs \citep{d'angelo2021repulsive,yashima22a}.
Comparing neural networks via a function kernel is also challenging since functions are infinite-dimensional objects.
Previous works resort to comparing functions only on a subset of the input space \citep{wang2018function,d'angelo2021repulsive}. Comparing functions over training data leads to underfitting \citep{d'angelo2021repulsive,yashima22a}, likely because these inputs have known labels, leaving no room for diverse predictions without impairing performance. 

\begin{figure}[t]
    \centering    
    \includegraphics[width=\textwidth]{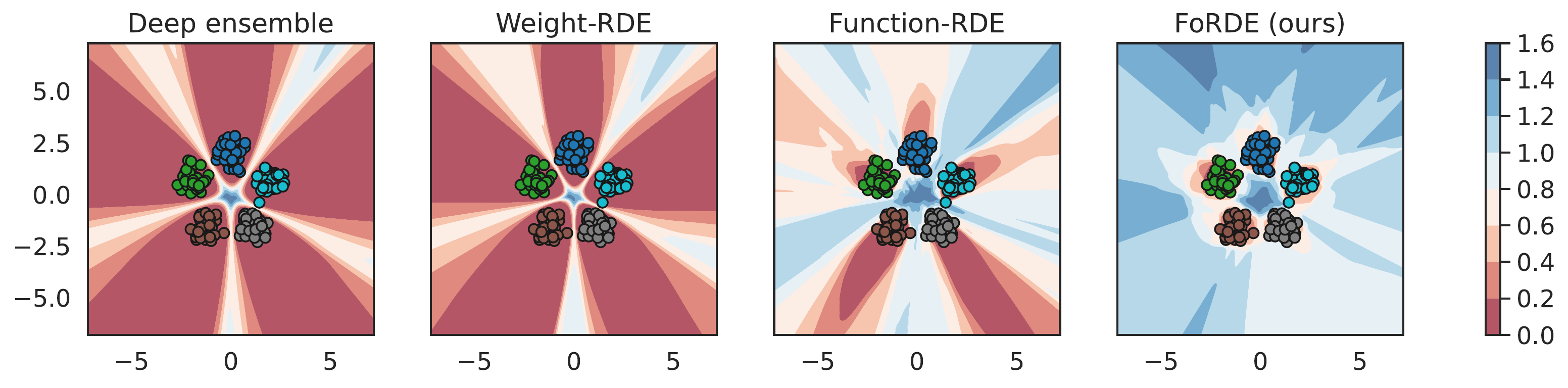}
    \caption{\textbf{Uncertainty of FoRDE is high in all input regions outside the training data, and is higher than baselines.} Each panel shows the entropy of the predictive posteriors in 2D classification.}
    \label{fig:2D_classification}
\end{figure}

\section{FoRDE: First-order Repulsive Deep Ensembles}

In this section, we present a framework to perform ParVI in the \emph{input-gradient} space.
We start by summarizing  Wasserstein gradient descent (WGD) in \cref{sec:wgd}, and show how to apply WGD for input-gradient-space repulsion in \cref{sec:wgd_for_input_grad}.
We then discuss hyperparameter selection for the input-gradient kernel in \cref{sec:pca_lengthscales}, and outline practical considerations in \cref{sec:practical_considerations}.  

Throughout this paper, we assume a set of $M$ weight particles $\{\theta_i\}_{i=1}^M$ corresponding to a set of $M$ neural networks $\{f_i: \x \mapsto f(\x; \theta_i)\}_{i=1}^M$.
We focus on the supervised classification setting: given a labelled dataset $\D=\{(\x_n, y_n)\}_{n=1}^N$ with $\mathcal{C}$ classes and inputs $\x_n \in \R^D$, we approximate the posterior $p(\theta | \D)$ using the $M$ particles. The output $f(\x; \theta)$ for input $\x$ is a vector of size $\mathcal{C}$ whose $y$-th entry $f(\x; \theta)_y$ is the logit of the $y$-th class.

\subsection{Wasserstein gradient descent}\label{sec:wgd}
Following \citet{d'angelo2021repulsive}, we use a ParVI method called Wasserstein gradient descent (WGD) \citep{liu2019povi,wang2022projected}.
Given an intractable target posterior distribution $\pi = p(\cdot | \D)$ and a set of particles $\{\theta_i\}_{i=1}^M$ from distribution $\rho$, the goal of WGD is to find the particle distribution $\rho^*$,
\begin{equation} \label{eq:wgd}
    \rho^* = \argmin_{\rho \in \mathcal{P}_2(\Theta)} \KL_{\pi}(\rho),
\end{equation}
where $\KL_{\pi}(\rho)$ is a shorthand for the standard Kullback-Leibler divergence
\begin{equation}
    \KL_{\pi}(\rho) = \E_{\rho(\theta)}\big[\log \rho(\theta)-\log \pi(\theta)\big],
\end{equation}
and $\mathcal{P}_2(\Theta)$ is the Wasserstein space equipped with the Wasserstein distance $W_2$ \citep{ambrosio2005gradient,villani2009optimal}.
WGD solves the problem in \cref{eq:wgd} using 
a Wasserstein gradient flow $(\rho_t)_t$, which is roughly the family of steepest descent curves of $\KL_{\pi}(\cdot)$ in $\mathcal{P}_2(\Theta)$.
The tangent vector of this gradient flow at time $t$ is
\begin{equation}\label{eq:wgd_tangent}
    v_t(\theta) = \nabla \log \pi(\theta) - \nabla \log \rho_t(\theta)
\end{equation}
whenever $\rho_t$ is absolutely continuous \citep{villani2009optimal,liu2019povi}.
Intuitively, $v_t(\theta)$ points to the direction where the probability mass at $\theta$ of $\rho_t$ should be transported in order to bring $\rho_t$ closer to $\pi$.
Since \cref{eq:wgd_tangent} requires the analytical form of $\rho_t$ which we do not have access to, we use kernel density estimation (KDE) to obtain a tractable approximation $\hat{\rho}_t$ induced by  particles $\{\theta_i^{(t)}\}_{i=1}^M$ at time $t$,
\begin{align}
    \hat{\rho}_t(\theta) \propto \sum_{i=1}^M k\big(\theta,\theta_i^{(t)}\big),
\end{align}
where $k$ is a positive semi-definite kernel. Then, the gradient of the approximation is
\begin{equation}\label{eq:kde_loggrad}
    \nabla \log \hat{\rho}_t(\theta) = \frac{\sum_{i=1}^M \nabla_{\theta} k\big(\theta,\theta_i^{(t)}\big)}{\sum_{i=1}^M k\big(\theta,\theta_i^{(t)}\big)}.
\end{equation}
Using \cref{eq:kde_loggrad} in \cref{eq:wgd_tangent}, we obtain a practical update rule for each particle $\theta^{(t)}$ of $\hat{\rho}_t$:
\begin{align}
    \theta^{(t+1)} 
               &= \theta^{(t)} + \eta_t \Bigg(\underbrace{\nabla_{\theta^{(t)}} \log \pi\big(\theta^{(t)}\big)}_{\text{driving force}} - \underbrace{\frac{\sum_{i=1}^M \nabla_{\theta^{(t)}} k\big(\theta^{(t)},\theta_i^{(t)}\big)}{\sum_{i=1}^M k\big(\theta^{(t)},\theta_i^{(t)}\big)}}_{\text{repulsion force}} \Bigg), \label{eq:wgd_kde}
\end{align}
where $\eta_t > 0$ is the step size at optimization time $t$. Intuitively, we can interpret the first term in the particle gradient as the driving force directing the particles towards high density regions of the posterior, while the second term is the repulsion force pushing the particles away from each other. 

\subsection{Defining the kernel for WGD in input gradient space}\label{sec:wgd_for_input_grad}



We propose to use a kernel comparing the \emph{input gradients} of the particles,
\begin{align}
    k(\theta_i, \theta_j) &\overset{\mathrm{def}}{=} \E_{(\x,y) \sim p(\x,y)}\Big[ \kappa\big(\nabla_\x f(\x; \theta_i)_{y}, \nabla_\x f(\x; \theta_j)_{y} \big) \Big],
\end{align}
where $\kappa$ is a \emph{base kernel} between gradients $\nabla_\x f(\x;\theta)_y$ that are of same size as the inputs $\x$. 
In essence, we define $k$ as the expected similarity between the input gradients of two networks with respect to the data distribution $p(\x,y)$. 
Interestingly, by using the kernel $k$, the KDE approximation $\hat{\rho}$ of the particle distribution not only depends on the particles themselves but also depends on the data distribution.
We approximate the kernel $k$ using the training samples, with linear complexity:
\begin{equation}\label{eq:input_grad_kernel}
    k(\theta_i, \theta_j) \approx k_{\D}(\theta_i, \theta_j) = \frac{1}{N}\sum_{n=1}^N \kappa\big(\nabla_\x f(\x_n; \theta_i)_{y_n}, \nabla_\x f(\x_n; \theta_j)_{y_n} \big).
\end{equation}
The kernel only compares the gradients of the true label $\nabla_\x f(\x_n; \theta)_{y_n}$, as opposed to the entire Jacobian matrix $\nabla_\x f(\x_n; \theta)$, as our motivation is to encourage each particle to learn different features that could explain the training sample $(\x_n,y_n)$ well.
This approach also reduces computational complexity, since automatic differentiation libraries such as JAX \citep{jax2018github} or Pytorch \citep{paszke2019pytorch} would require $\mathcal{C}$ passes, one per class, to calculate the full Jacobian.

\paragraph{Choosing the base kernel}
We choose the RBF kernel on the unit sphere as our base kernel $\kappa$:
\begin{align}\label{eq:base_kernel}
    \kappa(\s,\s'; \bS) = \exp\left( -\frac{1}{2} (\s-\s')^\top \bS^{-1} (\s-\s')\right), \qquad \s = \frac{\nabla_\x f(\x; \theta)_{y}}{||\nabla_\x f(\x; \theta)_{y}||_2} \in \R^D
\end{align}
where $\s,\s'$ denote the two normalized gradients of two particles with respect to one input, and $\bS \in \R^{D \times D}$ is a diagonal matrix containing squared lengthscales.
We design $\kappa$ to be norm-agnostic since the norm of the true label gradient $||\nabla_\x f(\x_n; \theta)_{y_n}||_2$ fluctuates during training and as training converges, the log-probability $f(\x_n; \theta)_{y_n}=\log p(y_n|\x_n,\theta)$ will approach $\log 1$, leading to the norm $||\nabla_\x f(\x_n; \theta)_{y_n}||_2$ approaching $0$ due to the saturation of the log-softmax activation. 
Furthermore, comparing the normed input gradients between ensemble members teaches them to learn complementary explanatory patterns from the training samples, which could improve robustness of the ensemble.
The RBF kernel is an apt kernel to compare unit vectors \citep{jayasumana2014optimizing}, and we can control the variances of the gradients along input dimensions via the square lengthscales $\bS$.


\subsection{Selecting the lengthscales for the base kernel} \label{sec:pca_lengthscales}
In this section, we present a method to select the lengthscales for the base kernel.
These lengthscales are important for the performance of FoRDE, since they control how much repulsion force is applied in each dimension of the input-gradient space. The dimension-wise repulsion is \citep{qiang2016svgd}
\begin{align}
    \frac{\partial}{\partial s_d} \kappa(\s,\s'; \bS) 
    &= -\frac{s_d-s'_d}{\bS_{dd}} \kappa(\s,\s'; \bS),
\end{align}
where we can see that along the $d$-th dimension the inverse square lengthscale $\bS_{dd}$ controls the strength of the repulsion $\nabla_{s_d} \kappa(\s,\s'; \bS)$: a smaller lengthscale corresponds to a stronger force.\footnote{Here we assume that the lengthscales are set appropriately so that the kernel $\kappa$ does not vanish, which is true since we use the median heuristic during training (\cref{sec:practical_considerations}).}
Additionally, since the repulsion is restricted to the unit sphere in the input-gradient space, increasing distance in one dimension decreases the distance in other dimensions.
As a result, the repulsion motivates the ensemble members to depend more on dimensions with stronger repulsion in the input gradient space for their predictions, while focusing less on dimensions with weaker repulsion. One should then apply stronger repulsion in dimensions of data manifold with higher variances.

To realize the intuition above in FoRDE, we first apply Principal Component Analysis (PCA) to discover dominant features in the data.
In PCA, we calculate the eigendecomposition
\begin{align}
    \mathbf{C} &= \U \bL \U^T \quad \in \R^{D \times D}
\end{align}
of the covariance matrix $\mathbf{C} = \mathbf{X}^T\mathbf{X}/(N-1)$ of the centered training samples $\mathbf{X} \in \R^{N \times D}$ to get eigenvectors and eigenvalues $\{\u_d, \lambda_d\}_{k=1}^D$. 
The $d$-th eigenvalue $\lambda_d$ is the variance of the data along eigenvector $\u_d$, offering a natural choice of inverse eigenvalues $\lambda_d^{-1}$ as the squared lengthscales $\bS_{dd}$ of the principal components. 
Let $\tilde{\x} = \U^T \x$ denote the representation of the input $\x$ in eigenbasis $\U = [\u_1 \, \u_2 \, \dots \, \u_D]$.
We compute the gradient kernel in PCA basis $\U$ and set $\bS = \bL^{-1}$: 
%
\begin{align}\label{eq:pca_kernel}
    \kappa(\s,\s') &\overset{\mathrm{def}}{=} 
    \exp\left(-\frac{1}{2} (\U^\top \s - \U^\top \s')^\top \,\bL\, (\U^\top \s - \U^\top \s')\right),
\end{align}
where $\bL$ is a diagonal eigenvalue matrix. 
While setting the square inverse lengthscales equal to the eigenvalues seems problematic at first glance, since large eigenvalues will push the kernel $\kappa$ towards 0, this problem is avoided in practice since we also employ the median heuristic, which introduces in the kernel a global bandwidth scaling term that adapts to the current pairwise distance between particles, as discussed below in \cref{sec:practical_considerations}.



\paragraph{Connection to the EmpCov prior} Recently, \citet{izmailov2021dangers} proposed the EmpCov prior for the weight columns $\mathbf{w}$ of the first layer of a BNN:
\begin{equation}
    \mathbf{w} \sim \mathcal{N}(0, \alpha \mathbf{C} + \epsilon \mathbf{I}), \qquad \mathbf{w} \in \R^D
\end{equation}
where $\alpha > 0$ determines the prior scale and $\epsilon > 0$ is a small constant ensuring a positive definite covariance.
The prior encourages first layer weights to vary more along higher variance data dimensions.  
Samples from this prior will have large input gradients along high variance input dimensions. 
In this sense, the EmpCov prior has a similar effect to the kernel in \cref{eq:pca_kernel} on ensemble members.
The difference is that while \cite{izmailov2021dangers} incorporates knowledge of the data manifold into the prior, we embed this knowledge into our approximate posterior via the kernel.

\subsection{Practical considerations} \label{sec:practical_considerations}

In this section, we detail two important considerations to make FoRDEs work in practice. We include the full training algorithm in \cref{sec:training_algo}.

\paragraph{Mini-batching} To make FoRDE amenable to mini-batch gradient optimization, we adapt the kernel in \cref{eq:input_grad_kernel} to a mini-batch of samples $\mathcal{B}=\{(\mathbf{x}_b, y_b)\}_{b=1}^B$:
\begin{equation}\label{eq:input_grad_minibatch_kernel}
    k_{\mathcal{B}}(\theta_i, \theta_j) = \frac{1}{B}\sum_{b=1}^B \kappa\Big(\nabla_\x f(\x_b; \theta_i)_{y_b}, \nabla_\x f(\x_b; \theta_j)_{y_b} \Big).
\end{equation}
In principle, this kernel in the update rule in \cref{eq:wgd_kde} leads to 
biased stochastic gradients of the repulsion term because the average over batch samples in \cref{eq:input_grad_minibatch_kernel} is inside the logarithm. However, in practice, we found no convergence issues in our experiments.

\paragraph{Median heuristics} Since we perform particle optimization with an RBF kernel $\kappa$, following earlier works \citep{qiang2016svgd,liu2019povi}, we adopt the median heuristic \citep{scholkopf2002learning}. Besides the lengthscales, we introduce a global bandwidth $h$ in our base kernel in \cref{eq:base_kernel}:
\begin{equation}\label{eq:median_heuristic}
    \kappa(\s_i,\s_j; \bS) = \exp\left( -\frac{1}{2h} (\s_i-\s_j)^\top \bS^{-1} (\s_i-\s_j)\right), \qquad \s_i = \frac{\nabla_\x f(\x; \theta_i)_{y}}{||\nabla_\x f(\x; \theta_i)_{y}||_2} \in \R^D.
\end{equation}
During training, the bandwidth $h$ is adaptively set to $\mathrm{med}^2/(2\log M)$, where $\mathrm{med}^2$ is the median of the pairwise distance $(\s_i-\s_j)^\top \bS^{-1} (\s_i-\s_j)$ between the weight samples $\{\theta_i\}_{i=1}^M$.

\subsection{Computational complexity}\label{sec:computational_complexity}
Compared to DEs, FoRDEs 
take roughly three times longer to train.
In addition to a forward-backward pass to calculate the log-likelihood, we need an additional forward-backward pass to calculate the input gradients, and another backward pass to calculate the gradients of the input gradient repulsion with respect to the weights.
This analysis is confirmed in practice: in \textsc{resnet18}/\textsc{cifar-100} experiments of \cref{sec:benchmarking} with an ensemble size of 10, a DE took $\sim$31 seconds per epoch on an Nvidia A100 GPU, while FoRDE took $\sim$101 seconds per epoch.

\section{Related works}

Besides the ParVI methods mentioned in \cref{sec:background}, we discuss additional related works below.

\paragraph{Diversifying input gradients of ensembles} Local independent training (LIT) \citep{ross2018learning} orthogonalizes the input gradients of ensemble members by minimizing their pairwise squared cosine similarities, and thus closely resembles our input-gradient repulsion term which diversifies input gradients on a hyper-sphere. However, their goal is to find a maximal set of models that accurately predict the data using different sets of distinct features, while our goal is to induce functional diversity in an ensemble. Furthermore, we formulate our kernelized repulsion term based on the ParVI framework, allowing us to choose hyperparameter settings (orthogonal basis and lengthscales) that imbue the ensemble with beneficial biases (such as robustness to corruption). 

\paragraph{Gradient-based attribution methods for deep models}
One application of input gradients is to build attribution (or \emph{saliency}) maps, which assign importance to visually-interpretable input features for a specified output \citep{simonyan2013deep,bach2015pixel,shrikumar2016not,sundararajan2017axiomatic,shrikumar2017learning}.
Our method intuitively utilizes the attribution perspective of input gradients to encourage ensemble members to learn different patterns from training data.

\paragraph{Improving corruption robustness of BNNs}
Previous works have evaluated the predictive uncertainty of BNNs under covariate shift \citep{ovadia2019can,izmailov21a}, with \citet{izmailov21a} showing that standard BNNs with high-fidelity posteriors perform worse than MAP solutions on under corruptions. \citet{izmailov2021dangers} attributed this phenomenon to the lack of posterior contraction in the null space of the data manifold and proposed the EmpCov prior as a remedy.
As discussed in \cref{sec:pca_lengthscales}, the PCA kernel works in the same manner as the EmpCov prior and thus significantly improves robustness of FoRDE against corruptions.
\citet{trinh22a} studied the robustness of node-BNNs, an efficient alternative to weight-based BNNs, and showed that by increasing the entropy of the posterior, node-BNNs become more robust against corruptions.
\cite{wang2023robustness} allow BNNs to adapt to the distribution shift at test time by using test data statistics.


\begin{table}[ht]
\centering
\caption{\textbf{FoRDE-PCA achieves the best performance under corruptions while FoRDE-Identity outperforms baselines on clean data. FoRDE-Tuned outperforms baselines on both clean and corrupted data.} Results of \textsc{resnet18} / \textsc{cifar-100} averaged over 5 seeds. Each ensemble has 10 members. cA, cNLL and cECE are accuracy, NLL, and ECE on \textsc{cifar-100-c}.}
\label{tab:cifar100_benchmark}
\resizebox{0.85\textwidth}{!}{%
\begin{sc}
\begin{tabular}{@{}lcccc@{}}
\toprule
Method         & NLL $\downarrow$ & Accuracy (\%) $\uparrow$ & ECE $\downarrow$  & cA / cNLL / cECE   \\ \midrule
node-BNNs     & $0.74 \textcolor{gray}{\pm 0.01}$  & $79.7 \textcolor{gray}{\pm 0.3} $          & $0.054 \textcolor{gray}{\pm 0.002}$ & 54.8 / 1.96 / 0.05 \\
SWAG           & $0.73 \textcolor{gray}{\pm 0.01}$  & $79.4 \textcolor{gray}{\pm 0.1}$           & $\mathbf{0.038 \textcolor{gray}{\pm 0.001}}$ & 53.0 / 2.03 / 0.05 \\ \midrule
Deep ensembles & $\mathbf{0.70 \textcolor{gray}{\pm 0.00}}$  & $81.8 \textcolor{gray}{\pm 0.2}$           & $0.041 \textcolor{gray}{\pm 0.003}$ & 54.3 / 1.99 / 0.05 \\
weight-RDE     & $\mathbf{0.70 \textcolor{gray}{\pm 0.01}}$  & $81.7 \textcolor{gray}{\pm 0.3}$           & $0.043 \textcolor{gray}{\pm 0.004}$ & 54.2 / 2.01 / 0.06 \\
function-RDE   & $0.76 \textcolor{gray}{\pm 0.02}$  & $80.1 \textcolor{gray}{\pm 0.4}$           & $0.042 \textcolor{gray}{\pm 0.005}$ & 51.9 / 2.08 / 0.07 \\
feature-RDE    & $0.75 \textcolor{gray}{\pm 0.04}$  & $\mathbf{82.1 \textcolor{gray}{\pm 0.3}}$           & $0.072 \textcolor{gray}{\pm 0.023}$ & 54.8 / 2.02 / 0.06 \\ \midrule
LIT & $\mathbf{0.70 \textcolor{gray}{\pm 0.00}}$  & $81.9 \textcolor{gray}{\pm 0.2}$           & $0.040 \textcolor{gray}{\pm 0.003}$ & 54.4 / 1.98 / 0.05 \\
FoRDE-PCA (ours)   & $0.71 \textcolor{gray}{\pm 0.00}$  & $81.4 \textcolor{gray}{\pm 0.2}$           & $0.039 \textcolor{gray}{\pm 0.002}$ & $\mathbf{56.1}$ / $\mathbf{1.90}$ / $\mathbf{0.05}$ \\
FoRDE-Identity (ours) & $\mathbf{0.70 \textcolor{gray}{\pm 0.00}}$ & $\mathbf{82.1 \textcolor{gray}{\pm 0.2}}$ & $0.043 \textcolor{gray}{\pm 0.001}$ & 54.1 / 2.02 / 0.05 \\
FoRDE-Tuned (ours) & $\mathbf{0.70 \textcolor{gray}{\pm 0.00}}$ & $\mathbf{82.1 \textcolor{gray}{\pm 0.2}}$ & $0.044 \textcolor{gray}{\pm 0.002}$ & 55.3 / 1.94 / 0.05 \\
\bottomrule
\end{tabular}
\end{sc}
}
\end{table}
\begin{table}[ht]
\centering
\caption{\textbf{FoRDE-PCA achieves the best performance under corruptions while FoRDE-Identity has the best NLL on clean data. FoRDE-Tuned outperforms most baselines on both clean and corrupted data.} Results of \textsc{resnet18} / \textsc{cifar-10} averaged over 5 seeds. Each ensemble has 10 members. cA, cNLL and cECE are accuracy, NLL, and ECE on \textsc{cifar-10-c}.}
\label{tab:cifar10_benchmark}
\resizebox{0.85\textwidth}{!}{%
\begin{sc}
\begin{tabular}{@{}lcccc@{}}
\toprule
Method         & NLL $\downarrow$ & Accuracy (\%) $\uparrow$ & ECE $\downarrow$  & cA / cNLL / cECE   \\ \midrule
node-BNNs      & $0.127 \textcolor{gray}{\pm 0.009}$ & $95.9 \textcolor{gray}{\pm 0.3}$           & $0.006 \textcolor{gray}{\pm 0.002}$ & 78.2 / 0.82 / 0.09 \\
SWAG           & $0.124 \textcolor{gray}{\pm 0.001}$ & $\mathbf{96.9 \textcolor{gray}{\pm 0.1}}$           & $0.005 \textcolor{gray}{\pm 0.001}$ & 77.5 / 0.78 / 0.07 \\ \midrule
Deep ensembles & $0.117 \textcolor{gray}{\pm 0.001}$ & $96.3 \textcolor{gray}{\pm 0.1}$           & $0.005 \textcolor{gray}{\pm 0.001}$ & 78.1 / 0.78 / 0.08 \\
weight-RDE     & $0.117 \textcolor{gray}{\pm 0.002}$ & $96.2 \textcolor{gray}{\pm 0.1}$           & $0.005 \textcolor{gray}{\pm 0.001}$ & 78.0 / 0.78 / 0.08 \\
function-RDE   & $0.128 \textcolor{gray}{\pm 0.001}$ & $95.8 \textcolor{gray}{\pm 0.2}$           & $0.006 \textcolor{gray}{\pm 0.001}$ & 77.1 / 0.81 / 0.08 \\
feature-RDE    & $0.116 \textcolor{gray}{\pm 0.001}$ & $96.4 \textcolor{gray}{\pm 0.1}$           & $\mathbf{0.004 \textcolor{gray}{\pm 0.001}}$ & 78.1 / 0.77 / 0.08 \\ \midrule
LIT & $0.116 \textcolor{gray}{\pm 0.001}$ & $96.4 \textcolor{gray}{\pm 0.1}$           & $0.004 \textcolor{gray}{\pm 0.001}$ & 78.2 / 0.78 / 0.09 \\
FoRDE-PCA (ours)   & $0.125 \textcolor{gray}{\pm 0.001}$ & $96.1 \textcolor{gray}{\pm 0.1}$           & $0.006 \textcolor{gray}{\pm 0.001}$ & $\mathbf{80.5}$ / $\mathbf{0.71}$ / $\mathbf{0.07}$ \\
FoRDE-Identity (ours) & $\mathbf{0.113 \textcolor{gray}{\pm 0.002}}$ & $96.3 \textcolor{gray}{\pm 0.1}$ & $0.005 \textcolor{gray}{\pm 0.001}$ & 78.0 / 0.80 / 0.08 \\
FoRDE-Tuned (ours) & $0.114 \textcolor{gray}{\pm 0.002}$ & $96.4 \textcolor{gray}{\pm 0.1}$ & $0.005 \textcolor{gray}{\pm 0.001}$ & 79.1 / 0.74 / 0.07 \\
\bottomrule
\end{tabular}
\end{sc}
}
\end{table}
\begin{table}[ht]
\centering
\caption{\textbf{FoRDE outperforms EmpCov priors under corruptions, while maintaining competitive performance on clean data.}
Results of \textsc{resnet18} on \textsc{cifar-10} evaluated over 5 seeds. Each ensemble has 10 members. cA, cNLL and cECE are accuracy, NLL, and ECE on \textsc{cifar-10-c}. Here we use the EmpCov prior for all methods except FoRDE.
}
\label{tab:cifar10_empcov}
\resizebox{0.85\textwidth}{!}{%
\begin{sc}
\begin{tabular}{@{}lcccc@{}}
\toprule
Method         & NLL $\downarrow$ & Accuracy (\%) $\uparrow$ & ECE $\downarrow$  & cA / cNLL / cECE   \\ \midrule
Deep ensembles & $0.119 \textcolor{gray}{\pm 0.001}$ & $\mathbf{96.2 \textcolor{gray}{\pm 0.1}}$ & $0.006 \textcolor{gray}{\pm 0.001}$ & 78.7 / 0.76 / 0.08 \\
weight-RDE     & $0.120 \textcolor{gray}{\pm 0.001}$ & $96.0 \textcolor{gray}{\pm 0.1}$ & $0.006 \textcolor{gray}{\pm 0.001}$ & 78.8 / 0.76 / 0.08 \\
function-RDE   & $0.132 \textcolor{gray}{\pm 0.001}$ & $95.6 \textcolor{gray}{\pm 0.3}$ & $0.007 \textcolor{gray}{\pm 0.001}$ & 77.8 / 0.79 / 0.08 \\
feature-RDE    & $\mathbf{0.118 \textcolor{gray}{\pm 0.001}}$ & $\mathbf{96.2 \textcolor{gray}{\pm 0.1}}$ & $\mathbf{0.005 \textcolor{gray}{\pm 0.001}}$ & 78.9 / 0.74 / 0.07 \\ \midrule
FoRDE-PCA (ours)   & $0.125 \textcolor{gray}{\pm 0.001}$ & $96.1 \textcolor{gray}{\pm 0.1}$           & $0.006 \textcolor{gray}{\pm 0.001}$ & $\mathbf{80.5}$ / $\mathbf{0.71}$ / $\mathbf{0.07}$ \\
\bottomrule
\end{tabular}
\end{sc}
}
\end{table}

\section{Experiments}
\subsection{Illustrating functional diversity}
To show that FoRDE does produce better functional diversity than plain DE and other repulsive DE approaches, we repeated the 1D regression of \citet{izmailov_subspace_2019} and the 2D classification experiments of \citet{d'angelo2021repulsive}.
We use ensembles of 16 networks for these experiments.
\cref{fig:1D_regression} shows that FoRDE exhibits higher predictive uncertainty in the input regions outside the training data compared to the baselines in 1D regression.
For the 2D classification task, we visualize the entropy of the predictive posteriors in \cref{fig:2D_classification}, which also shows that FoRDE has higher uncertainty than the baselines. Furthermore, FoRDE is the only method that exhibits high uncertainty in all input regions outside the training data, a property mainly observed in predictive uncertainty of Gaussian processes \citep{rasmussen_gp}.

\subsection{Comparisons to other repulsive DE methods and BNNs} \label{sec:benchmarking}

We report performance of FoRDE against other methods on \textsc{cifar-10/100} \citep{krizhevsky2009cifar} in \crefrange{tab:cifar100_benchmark}{tab:cifar10_benchmark} and \textsc{tinyimagenet} \citep{Le2015TinyIV} in \cref{sec:tinyimagenet_results}.
Besides the PCA lengthscales introduced in \cref{sec:pca_lengthscales}, we experiment with the identity lengthscales $\boldsymbol{\Sigma}=\mathbf{I}$ and with tuned lengthscales where we take the weighted average of the PCA lengthscales and the identity lengthscales.
Details on lengthscale tuning are presented in \cref{sec:tuning_lengthscales}.
For the repulsive DE (RDE) baselines, we choose weight RDE \citep{d'angelo2021repulsive}, function RDE \citep{d'angelo2021repulsive} and feature RDE \citep{yashima22a}.
We also include LIT \citep{ross2018learning}, node-BNNs \citep{trinh22a} and SWAG \citep{maddox2019} as baselines.
We use an ensemble size of 10.
We use standard performance metrics of expected calibration error (ECE) \citep{Naeini2015}, negative log-likelihood (NLL) and predictive accuracy.
For evaluations on input perturbations, we use \textsc{cifar-10/100-c} and \textsc{tinyimagenet-c} provided by \citet{hendrycks2016baseline}, which are datasets of corrupted test images containing 19 image corruption types across 5 levels of severity, and we report the accuracy, NLL and ECE averaged over all corruption types and severity levels (denoted cA, cNLL and cECE in \crefrange{tab:cifar100_benchmark}{tab:tinyimagenet_benchmark}).
We use \textsc{resnet18}  \citep{he2016deep} for \textsc{cifar-10/100} and \textsc{preactresnet18} \citep{he2016identity} for \textsc{tinyimagenet}.
Experimental details are included in \cref{sec:image_exp_details}.

\cref{tab:cifar100_benchmark,tab:cifar10_benchmark} show that FoRDE-PCA outperforms other methods under input corruptions across all metrics, while maintaining competitive performance on clean data.
For instance, FoRDE-PCA shows a $+1.3\%$ gain on \textsc{cifar-100-c} and $+2.4\%$ gain on \textsc{cifar-10-c} in accuracy compared to the second-best results.
As stated in \cref{sec:pca_lengthscales}, the PCA kernel encourages FoRDE to rely more on features with high variances in the data manifold to make predictions, while being less dependent on features with low variances. This effect has been shown in \citet{izmailov2021dangers} to boost model robustness against perturbations, which explains why FoRDE with the PCA kernel performs better than the baselines on input corruptions.

On the other hand, \cref{tab:cifar100_benchmark,tab:cifar10_benchmark} show that FoRDE with identity lengthscales outperforms the baselines in terms of NLL on \textsc{cifar-10} and has the best accuracy on \textsc{cifar-100}. However, FoRDE-Identity is slightly less robust than DE against corruptions. We suspect that with the identity lengthscales, FoRDE also learns to rely on low-variance features to make predictions, which is harmful to performance under corruptions \citep{izmailov2021dangers}.

Finally, \cref{tab:cifar100_benchmark,tab:cifar10_benchmark} show that FoRDE with tuned lengthscales (FoRDE-Tuned) outperforms the baselines on both clean and corrupted data in most cases, suggesting that the optimal lengthscales for good performance on both clean and corrupted data lie somewhere between the identity lengthscales and the PCA lengthscales. Additional results on lengthscale tuning are presented in \cref{sec:tuning_lengthscales}.

\begin{figure}[t]
    \centering
    \includegraphics[width=\textwidth]{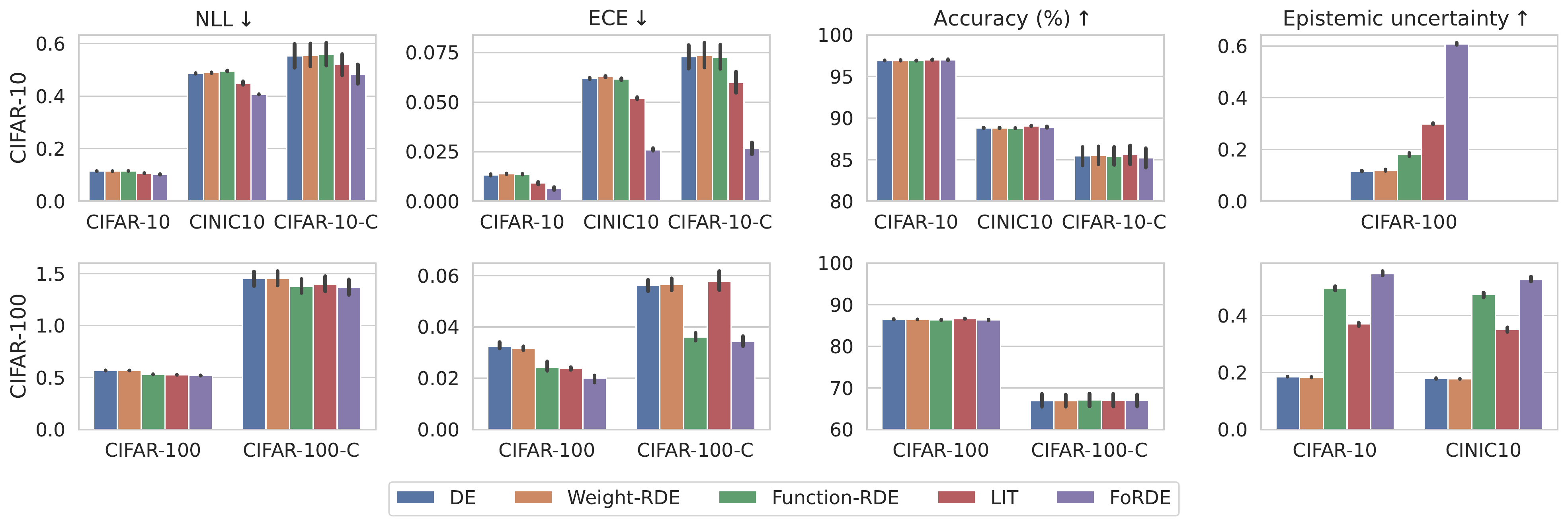}
    \caption{\textbf{FoRDE outperforms competing methods in transfer learning.} 
    \textbf{First three columns:} We report NLL, ECE and accuracy on in-distribution test sets and under covariate shift. For \textsc{cifar-10}, we use \textsc{cinic10} \citep{darlow2018cinic} to evaluate models under natural shift and \textsc{cifar-10-c} for corruption shift. For \textsc{cifar-100}, we evaluate on \textsc{cifar-100-c}. FoRDE performs better than the baselines in all cases. \textbf{Last column:} We evaluate functional diversity by calculating epistemic uncertainty of ensembles on out-of-distribution (OOD) datasets using the formula in \citet{pmlr-v80-depeweg18a}. We use \textsc{cifar-100} as the OOD test set for \textsc{cifar-10} and we use \textsc{cifar-10} and \textsc{cinic10} as OOD test sets for \textsc{cifar-100}. FoRDE exhibits higher functional diversity than the baselines.}
    \label{fig:cifar_transferlearning}
\end{figure}
\subsection{Comparisons to EmpCov prior}
As stated in \cref{sec:pca_lengthscales}, our approach is similar to the EmpCov prior \citep{izmailov2021dangers}. We thus perform comparisons against ensemble methods where the EmpCov prior is defined for the first layer instead of the standard isotropic Gaussian prior.
We report the results of \textsc{resnet18}/\textsc{cifar-10} in \cref{tab:cifar10_empcov}, where we use the EmpCov prior for all ensemble methods except FoRDE.
Comparing \cref{tab:cifar10_empcov} to \cref{tab:cifar10_benchmark} indicates that the EmpCov prior slightly improves robustness of the baseline ensemble methods against corruptions, while also leading to a small reduction in performance on clean data.
These small improvements in robustness are not surprising, since for ensemble methods consisting of approximate MAP solutions, the isotropic Gaussian prior already minimizes the influences of low-variance data features on the ensemble's predictions \citep{izmailov2021dangers}.
We argue that besides minimizing the influences of low variance features on predictions via the PCA kernel, FoRDE also encourages its members to learn complementary patterns that can explain the data well, and these two effects act in synergy to improve the robustness of the resulting ensemble.
Thus, \cref{tab:cifar10_empcov} shows that FoRDE is still more robust against corruptions than the baseline methods with the EmpCov prior.

\subsection{Transfer learning experiments}
To show the practicality of FoRDE, we evaluated its performance in a transfer learning scenario. We use the outputs of the last hidden layer of a Vision Transformer model pretrained on \textsc{imagenet-21k} as input features and train ensembles of 10 networks. We report the results on \textsc{cifar-10} in the first row and on \textsc{cifar-100} in the second row in \cref{fig:cifar_transferlearning}.
Overall, \cref{fig:cifar_transferlearning} shows that FoRDE is better than the baselines across all cases. See \cref{sec:transferlearning_details} for experimental details.

\section{Discussion}\label{sec:discussion}
In this section, we outline directions to further improve FoRDE.

\paragraph{Reducing computational complexity} One major drawback of FoRDEs is the high computational complexity as discussed in \cref{sec:computational_complexity}. To circumvent this problem, one could either (i) only calculate the repulsion term after every $k > 1$ epochs, or (ii) using only a subset of batch samples at each iteration to calculate the repulsion term.

\paragraph{Reducing the number of lengthscale parameters} Here we use the RBF kernel as our base kernel, which requires us to choose appropriate lengthscales for good performance. To avoid this problem, we could explore other kernels suitable for unit vector comparisons, such as those introduced in \cite{jayasumana2014optimizing}. Another solution is to study dimensionality reduction techniques for input gradients before calculating the kernel, which can reduce the number of lengthscales to be set.


\section{Conclusion}
In this work, we proposed FoRDE, an ensemble learning method that promotes diversity in the input-gradient space among ensemble members.
We detailed the update rule and devised a data-dependent kernel suitable for input-gradient repulsion.
Experiments on image classification and transfer learning tasks show that FoRDE outperforms other ensemble methods under covariate shift.
Future directions include more efficient implementations of the method and reducing the burden of hyperparameter selection as discussed in \cref{sec:discussion}.

\subsubsection*{Acknowledgments}
This work was supported by the Research Council of Finland (Flagship programme: Finnish Center for Artificial Intelligence FCAI and decision no. 359567, 345604 and 341763), ELISE Networks of Excellence Centres (EU Horizon: 2020 grant agreement 951847) and UKRI Turing AI World-Leading Researcher Fellowship (EP/W002973/1). We acknowledge the computational resources provided by Aalto Science-IT project and CSC–IT Center for Science, Finland.

\section*{Ethics statement}
Our paper introduces a new ensemble learning method for neural networks, allowing deep learning models to be more reliable in practice.
Therefore, we believe that our work contributes towards making neural networks safer and more reliable to use in real-world applications, especially those that are safety-critical.
Our technique per se does not directly deal with issues such as fairness, bias or other potentially harmful societal impacts, which may be caused by improper usages of machine learning or deep learning systems \citep{aifairness}.
These issues would need to be adequately considered when constructing the datasets and designing specific deep learning applications.

\section*{Reproducibility statement}
For the purpose of reproducibility of our results with our new ensemble learning method, we have included in the Appendix detailed descriptions of the training algorithm.
For each experiment, we include in the Appendix details about the neural network architecture, datasets, data augmentation procedures and hyperparameter settings.
All datasets used for our experiments are publicly available.
We have included our codes in the supplementary material and we provide instructions on how to run our experiments in a \texttt{README.md} available in the provided codebase.
For the transfer learning experiments, we used publicly available pretrained models which we have mentioned in the Appendix.

\bibliography{refs}

\begin{thebibliography}{53}
\providecommand{\natexlab}[1]{#1}
\providecommand{\url}[1]{\texttt{#1}}
\expandafter\ifx\csname urlstyle\endcsname\relax
  \providecommand{\doi}[1]{doi: #1}\else
  \providecommand{\doi}{doi: \begingroup \urlstyle{rm}\Url}\fi

\bibitem[Ambrosio et~al.(2005)Ambrosio, Gigli, and
  Savar{\'e}]{ambrosio2005gradient}
Luigi Ambrosio, Nicola Gigli, and Giuseppe Savar{\'e}.
\newblock \emph{Gradient flows: in metric spaces and in the space of
  probability measures}.
\newblock Springer, 2005.

\bibitem[Ashukha et~al.(2020)Ashukha, Lyzhov, Molchanov, and
  Vetrov]{Ashukha2020Pitfalls}
Arsenii Ashukha, Alexander Lyzhov, Dmitry Molchanov, and Dmitry Vetrov.
\newblock Pitfalls of in-domain uncertainty estimation and ensembling in deep
  learning.
\newblock In \emph{International Conference on Learning Representations}, 2020.

\bibitem[Bach et~al.(2015)Bach, Binder, Montavon, Klauschen, M{\"u}ller, and
  Samek]{bach2015pixel}
Sebastian Bach, Alexander Binder, Gr{\'e}goire Montavon, Frederick Klauschen,
  Klaus-Robert M{\"u}ller, and Wojciech Samek.
\newblock On pixel-wise explanations for non-linear classifier decisions by
  layer-wise relevance propagation.
\newblock \emph{PloS ONE}, 10\penalty0 (7):\penalty0 e0130140, 2015.

\bibitem[Blundell et~al.(2015)Blundell, Cornebise, Kavukcuoglu, and
  Wierstra]{blundell2015weight}
Charles Blundell, Julien Cornebise, Koray Kavukcuoglu, and Daan Wierstra.
\newblock Weight uncertainty in neural network.
\newblock In \emph{International conference on machine learning}, 2015.

\bibitem[Bradbury et~al.(2018)Bradbury, Frostig, Hawkins, Johnson, Leary,
  Maclaurin, Necula, Paszke, Vander{P}las, Wanderman-{M}ilne, and
  Zhang]{jax2018github}
James Bradbury, Roy Frostig, Peter Hawkins, Matthew~James Johnson, Chris Leary,
  Dougal Maclaurin, George Necula, Adam Paszke, Jake Vander{P}las, Skye
  Wanderman-{M}ilne, and Qiao Zhang.
\newblock {JAX}: composable transformations of {P}ython+{N}um{P}y programs,
  2018.

\bibitem[Chen et~al.(2018)Chen, Zhang, Wang, Li, and Chen]{chen2018unified}
Changyou Chen, Ruiyi Zhang, Wenlin Wang, Bai Li, and Liqun Chen.
\newblock A unified particle-optimization framework for scalable {B}ayesian
  sampling.
\newblock In \emph{Uncertainty in Artificial Intelligence}, 2018.

\bibitem[D'Angelo \& Fortuin(2021)D'Angelo and Fortuin]{d'angelo2021repulsive}
Francesco D'Angelo and Vincent Fortuin.
\newblock Repulsive deep ensembles are {B}ayesian.
\newblock In \emph{Advances in Neural Information Processing Systems}, 2021.

\bibitem[D'Angelo et~al.(2021)D'Angelo, Fortuin, and Wenzel]{d2021stein}
Francesco D'Angelo, Vincent Fortuin, and Florian Wenzel.
\newblock On {S}tein variational neural network ensembles.
\newblock In \emph{ICML workshop Uncertainty and Robustness in Deep Learning},
  2021.

\bibitem[Darlow et~al.(2018)Darlow, Crowley, Antoniou, and
  Storkey]{darlow2018cinic}
Luke~N Darlow, Elliot~J Crowley, Antreas Antoniou, and Amos~J Storkey.
\newblock Cinic-10 is not imagenet or cifar-10.
\newblock \emph{arXiv preprint arXiv:1810.03505}, 2018.

\bibitem[Depeweg et~al.(2018)Depeweg, Hernandez-Lobato, Doshi-Velez, and
  Udluft]{pmlr-v80-depeweg18a}
Stefan Depeweg, Jose-Miguel Hernandez-Lobato, Finale Doshi-Velez, and Steffen
  Udluft.
\newblock Decomposition of uncertainty in {B}ayesian deep learning for
  efficient and risk-sensitive learning.
\newblock In Jennifer Dy and Andreas Krause (eds.), \emph{Proceedings of the
  35th International Conference on Machine Learning}, volume~80 of
  \emph{Proceedings of Machine Learning Research}, pp.\  1184--1193. PMLR,
  10--15 Jul 2018.
\newblock URL \url{https://proceedings.mlr.press/v80/depeweg18a.html}.

\bibitem[Dietterich(2000)]{emsembleinML}
Thomas~G. Dietterich.
\newblock Ensemble methods in machine learning.
\newblock In \emph{International Workshop on Multiple Classifier Systems},
  2000.

\bibitem[Entezari et~al.(2022)Entezari, Sedghi, Saukh, and
  Neyshabur]{entezari2022the}
Rahim Entezari, Hanie Sedghi, Olga Saukh, and Behnam Neyshabur.
\newblock The role of permutation invariance in linear mode connectivity of
  neural networks.
\newblock In \emph{International Conference on Learning Representations}, 2022.

\bibitem[Fort et~al.(2019)Fort, Hu, and Lakshminarayanan]{fort2019deep}
Stanislav Fort, Huiyi Hu, and Balaji Lakshminarayanan.
\newblock Deep ensembles: A loss landscape perspective.
\newblock In \emph{NeurIPS workshop Bayesian Deep Learning}, 2019.

\bibitem[Graves(2011)]{graves2011practical}
Alex Graves.
\newblock Practical variational inference for neural networks.
\newblock In \emph{Advances in Neural Information Processing Systems}, 2011.

\bibitem[Gustafsson et~al.(2020)Gustafsson, Danelljan, and
  Schon]{gustafsson2020evaluating}
Fredrik~K Gustafsson, Martin Danelljan, and Thomas~B Schon.
\newblock Evaluating scalable bayesian deep learning methods for robust
  computer vision.
\newblock In \emph{IEEE/CVF Conference on Computer Vision and Pattern
  Recognition workshops}, 2020.

\bibitem[He et~al.(2016{\natexlab{a}})He, Zhang, Ren, and Sun]{he2016deep}
Kaiming He, Xiangyu Zhang, Shaoqing Ren, and Jian Sun.
\newblock Deep residual learning for image recognition.
\newblock In \emph{IEEE conference on Computer Vision and Pattern Recognition},
  2016{\natexlab{a}}.

\bibitem[He et~al.(2016{\natexlab{b}})He, Zhang, Ren, and Sun]{he2016identity}
Kaiming He, Xiangyu Zhang, Shaoqing Ren, and Jian Sun.
\newblock Identity mappings in deep residual networks.
\newblock In \emph{European Conference on Computer Vision}, 2016{\natexlab{b}}.

\bibitem[Hendrycks \& Dietterich(2019)Hendrycks and
  Dietterich]{hendrycks2018benchmarking}
Dan Hendrycks and Thomas Dietterich.
\newblock Benchmarking neural network robustness to common corruptions and
  perturbations.
\newblock In \emph{International Conference on Learning Representations}, 2019.

\bibitem[Hendrycks \& Gimpel(2017)Hendrycks and Gimpel]{hendrycks2016baseline}
Dan Hendrycks and Kevin Gimpel.
\newblock A baseline for detecting misclassified and out-of-distribution
  examples in neural networks.
\newblock In \emph{International Conference on Learning Representations}, 2017.

\bibitem[Izmailov et~al.(2019)Izmailov, Maddox, Kirichenko, Garipov, Vetrov,
  and Wilson]{izmailov_subspace_2019}
Pavel Izmailov, Wesley Maddox, Polina Kirichenko, Timur Garipov, Dmitry Vetrov,
  and Andrew~Gordon Wilson.
\newblock Subspace inference for bayesian deep learning.
\newblock \emph{Uncertainty in Artificial Intelligence (UAI)}, 2019.

\bibitem[Izmailov et~al.(2021{\natexlab{a}})Izmailov, Nicholson, Lotfi, and
  Wilson]{izmailov2021dangers}
Pavel Izmailov, Patrick Nicholson, Sanae Lotfi, and Andrew~G Wilson.
\newblock Dangers of {Bayesian} model averaging under covariate shift.
\newblock In \emph{Advances in Neural Information Processing Systems},
  2021{\natexlab{a}}.

\bibitem[Izmailov et~al.(2021{\natexlab{b}})Izmailov, Vikram, Hoffman, and
  Wilson]{izmailov21a}
Pavel Izmailov, Sharad Vikram, Matthew~D Hoffman, and Andrew~Gordon Wilson.
\newblock What are bayesian neural network posteriors really like?
\newblock In \emph{International Conference on Machine Learning},
  2021{\natexlab{b}}.

\bibitem[Jayasumana et~al.(2014)Jayasumana, Hartley, Salzmann, Li, and
  Harandi]{jayasumana2014optimizing}
Sadeep Jayasumana, Richard Hartley, Mathieu Salzmann, Hongdong Li, and Mehrtash
  Harandi.
\newblock Optimizing over radial kernels on compact manifolds.
\newblock In \emph{IEEE Conference on Computer Vision and Pattern Recognition},
  2014.

\bibitem[Krizhevsky(2009)]{krizhevsky2009cifar}
Alex Krizhevsky.
\newblock Learning multiple layers of features from tiny images.
\newblock Technical report, 2009.

\bibitem[Lakshminarayanan et~al.(2017)Lakshminarayanan, Pritzel, and
  Blundell]{lakshminarayanan2017simple}
Balaji Lakshminarayanan, Alexander Pritzel, and Charles Blundell.
\newblock Simple and scalable predictive uncertainty estimation using deep
  ensembles.
\newblock In \emph{Advances in Neural Information Processing Systems}, 2017.

\bibitem[Le \& Yang(2015)Le and Yang]{Le2015TinyIV}
Ya~Le and Xuan~S. Yang.
\newblock Tiny {ImageNet} visual recognition challenge.
\newblock 2015.

\bibitem[Liu et~al.(2019)Liu, Zhuo, Cheng, Zhang, and Zhu]{liu2019povi}
Chang Liu, Jingwei Zhuo, Pengyu Cheng, Ruiyi Zhang, and Jun Zhu.
\newblock Understanding and accelerating particle-based variational inference.
\newblock In \emph{International Conference on Machine Learning}, 2019.

\bibitem[Liu \& Wang(2016)Liu and Wang]{qiang2016svgd}
Qiang Liu and Dilin Wang.
\newblock Stein variational gradient descent: A general purpose {B}ayesian
  inference algorithm.
\newblock In \emph{Advances in Neural Information Processing Systems}, 2016.

\bibitem[Maddox et~al.(2019)Maddox, Izmailov, Garipov, Vetrov, and
  Wilson]{maddox2019}
Wesley~J Maddox, Pavel Izmailov, Timur Garipov, Dmitry~P Vetrov, and
  Andrew~Gordon Wilson.
\newblock A simple baseline for {B}ayesian uncertainty in deep learning.
\newblock In \emph{Advances in Neural Information Processing Systems}, 2019.

\bibitem[Mehrabi et~al.(2021)Mehrabi, Morstatter, Saxena, Lerman, and
  Galstyan]{aifairness}
Ninareh Mehrabi, Fred Morstatter, Nripsuta Saxena, Kristina Lerman, and Aram
  Galstyan.
\newblock A survey on bias and fairness in machine learning.
\newblock \emph{ACM Comput. Surv.}, 54\penalty0 (6), jul 2021.
\newblock ISSN 0360-0300.
\newblock \doi{10.1145/3457607}.
\newblock URL \url{https://doi.org/10.1145/3457607}.

\bibitem[Naeini et~al.(2015)Naeini, Cooper, and Hauskrecht]{Naeini2015}
Mahdi~Pakdaman Naeini, Gregory~F. Cooper, and Milos Hauskrecht.
\newblock Obtaining well calibrated probabilities using {B}ayesian binning.
\newblock In \emph{AAAI Conference on Artificial Intelligence}, 2015.

\bibitem[Neal(2012)]{neal2012bayesian}
Radford~M Neal.
\newblock \emph{Bayesian learning for neural networks}, volume 118 of
  \emph{Lecture Notes in Statistics}.
\newblock Springer, 2012.

\bibitem[Ovadia et~al.(2019)Ovadia, Fertig, Ren, Nado, Sculley, Nowozin,
  Dillon, Lakshminarayanan, and Snoek]{ovadia2019can}
Yaniv Ovadia, Emily Fertig, Jie Ren, Zachary Nado, David Sculley, Sebastian
  Nowozin, Joshua Dillon, Balaji Lakshminarayanan, and Jasper Snoek.
\newblock Can you trust your model's uncertainty? {E}valuating predictive
  uncertainty under dataset shift.
\newblock In \emph{Advances in Neural Information Processing Systems}, 2019.

\bibitem[Pang et~al.(2019)Pang, Xu, Du, Chen, and Zhu]{pmlr-v97-pang19a}
Tianyu Pang, Kun Xu, Chao Du, Ning Chen, and Jun Zhu.
\newblock Improving adversarial robustness via promoting ensemble diversity.
\newblock In \emph{International Conference on Machine Learning}, 2019.

\bibitem[Paszke et~al.(2019)Paszke, Gross, Massa, Lerer, Bradbury, Chanan,
  Killeen, Lin, Gimelshein, Antiga, Desmaison, Kopf, Yang, DeVito, Raison,
  Tejani, Chilamkurthy, Steiner, Fang, Bai, and Chintala]{paszke2019pytorch}
Adam Paszke, Sam Gross, Francisco Massa, Adam Lerer, James Bradbury, Gregory
  Chanan, Trevor Killeen, Zeming Lin, Natalia Gimelshein, Luca Antiga, Alban
  Desmaison, Andreas Kopf, Edward Yang, Zachary DeVito, Martin Raison, Alykhan
  Tejani, Sasank Chilamkurthy, Benoit Steiner, Lu~Fang, Junjie Bai, and Soumith
  Chintala.
\newblock Pytorch: An imperative style, high-performance deep learning library.
\newblock In \emph{Advances in Neural Information Processing Systems}, 2019.

\bibitem[Rame \& Cord(2021)Rame and Cord]{rame2021dice}
Alexandre Rame and Matthieu Cord.
\newblock {{DICE}}: Diversity in deep ensembles via conditional redundancy
  adversarial estimation.
\newblock In \emph{International Conference on Learning Representations}, 2021.

\bibitem[Rasmussen \& Williams(2006)Rasmussen and Williams]{rasmussen_gp}
Carl~Edward Rasmussen and Christopher K.~I. Williams.
\newblock \emph{Gaussian processes for machine learning.}
\newblock Adaptive computation and machine learning. MIT Press, 2006.
\newblock ISBN 026218253X.

\bibitem[Ross et~al.(2018)Ross, Pan, and Doshi-Velez]{ross2018learning}
Andrew~Slavin Ross, Weiwei Pan, and Finale Doshi-Velez.
\newblock Learning qualitatively diverse and interpretable rules for
  classification.
\newblock \emph{arXiv preprint arXiv:1806.08716}, 2018.

\bibitem[Sch{\"o}lkopf et~al.(2002)Sch{\"o}lkopf, Smola, Bach,
  et~al.]{scholkopf2002learning}
Bernhard Sch{\"o}lkopf, Alexander~J Smola, Francis Bach, et~al.
\newblock \emph{Learning with kernels: support vector machines, regularization,
  optimization, and beyond}.
\newblock MIT press, 2002.

\bibitem[Shen et~al.(2021)Shen, Heinonen, and Kaski]{shen2021}
Zheyang Shen, Markus Heinonen, and Samuel Kaski.
\newblock De-randomizing {MCMC} dynamics with the diffusion {Stein} operator.
\newblock In \emph{Advances in {Neural} {Information} {Processing} {Systems}},
  2021.

\bibitem[Shrikumar et~al.(2016)Shrikumar, Greenside, Shcherbina, and
  Kundaje]{shrikumar2016not}
Avanti Shrikumar, Peyton Greenside, Anna Shcherbina, and Anshul Kundaje.
\newblock Not just a black box: Learning important features through propagating
  activation differences.
\newblock \emph{arXiv}, 2016.

\bibitem[Shrikumar et~al.(2017)Shrikumar, Greenside, and
  Kundaje]{shrikumar2017learning}
Avanti Shrikumar, Peyton Greenside, and Anshul Kundaje.
\newblock Learning important features through propagating activation
  differences.
\newblock In \emph{International conference on machine learning}, 2017.

\bibitem[Simonyan et~al.(2014)Simonyan, Vedaldi, and
  Zisserman]{simonyan2013deep}
Karen Simonyan, Andrea Vedaldi, and Andrew Zisserman.
\newblock Deep inside convolutional networks: Visualising image classification
  models and saliency maps.
\newblock In \emph{International Conference on Learning Representations
  Workshop}, 2014.

\bibitem[Sundararajan et~al.(2017)Sundararajan, Taly, and
  Yan]{sundararajan2017axiomatic}
Mukund Sundararajan, Ankur Taly, and Qiqi Yan.
\newblock Axiomatic attribution for deep networks.
\newblock In \emph{International conference on machine learning}, 2017.

\bibitem[Trinh et~al.(2022)Trinh, Heinonen, Acerbi, and Kaski]{trinh22a}
Trung~Q Trinh, Markus Heinonen, Luigi Acerbi, and Samuel Kaski.
\newblock Tackling covariate shift with node-based {B}ayesian neural networks.
\newblock In \emph{International Conference on Machine Learning}, 2022.

\bibitem[Villani(2009)]{villani2009optimal}
C{\'e}dric Villani.
\newblock \emph{Optimal transport: old and new}, volume 338 of
  \emph{Grundlehren der mathematischen Wissenschaften}.
\newblock Springer, 2009.

\bibitem[Wang \& Aitchison(2023)Wang and Aitchison]{wang2023robustness}
Xi~Wang and Laurence Aitchison.
\newblock Robustness to corruption in pre-trained {B}ayesian neural networks.
\newblock In \emph{International Conference on Learning Representations}, 2023.

\bibitem[Wang et~al.(2022)Wang, Chen, and Li]{wang2022projected}
Yifei Wang, Peng Chen, and Wuchen Li.
\newblock Projected {Wasserstein} gradient descent for high-dimensional
  {B}ayesian inference.
\newblock \emph{SIAM/ASA Journal on Uncertainty Quantification}, 10\penalty0
  (4):\penalty0 1513--1532, 2022.

\bibitem[Wang et~al.(2019)Wang, Ren, Zhu, and Zhang]{wang2018function}
Ziyu Wang, Tongzheng Ren, Jun Zhu, and Bo~Zhang.
\newblock Function space particle optimization for {B}ayesian neural networks.
\newblock In \emph{International Conference on Learning Representations}, 2019.

\bibitem[Welling \& Teh(2011)Welling and Teh]{welling2011bayesian}
Max Welling and Yee~W Teh.
\newblock Bayesian learning via stochastic gradient {L}angevin dynamics.
\newblock In \emph{International Conference on Machine Learning}, 2011.

\bibitem[Yashima et~al.(2022)Yashima, Suzuki, Ishikawa, Sato, and
  Kawakami]{yashima22a}
Shingo Yashima, Teppei Suzuki, Kohta Ishikawa, Ikuro Sato, and Rei Kawakami.
\newblock Feature space particle inference for neural network ensembles.
\newblock In \emph{International Conference on Machine Learning}, 2022.

\bibitem[Zagoruyko \& Komodakis(2016)Zagoruyko and
  Komodakis]{zagoruyko2016wide}
Sergey Zagoruyko and Nikos Komodakis.
\newblock Wide residual networks.
\newblock In \emph{British Machine Vision Conference}, 2016.

\bibitem[Zhang et~al.(2020)Zhang, Li, Zhang, Chen, and
  Wilson]{Zhang2020Cyclical}
Ruqi Zhang, Chunyuan Li, Jianyi Zhang, Changyou Chen, and Andrew~Gordon Wilson.
\newblock Cyclical stochastic gradient {MCMC} for {B}ayesian deep learning.
\newblock In \emph{International Conference on Learning Representations}, 2020.

\end{thebibliography}
\bibliographystyle{iclr2024_conference}

\newpage
\appendix

\section{Results on \textsc{tinyimagenet}}\label{sec:tinyimagenet_results}
\begin{table}[h]
\centering
\caption{\textbf{FoRDE-PCA performs best under corruptions while having competitive performance on clean data.} Results of \textsc{preactresnet18} on \textsc{tinyimagenet} evaluated over 5 seeds. Each ensemble has 10 members. cA, cNLL and cECE are accuracy, NLL, and ECE on \textsc{tinyimagenet-c}.}
\label{tab:tinyimagenet_benchmark}
\resizebox{0.9\textwidth}{!}{%
\begin{sc}
\begin{tabular}{@{}lcccc@{}}
\toprule
Method         & NLL $\downarrow$ & Accuracy (\%) $\uparrow$ & ECE $\downarrow$  & cA / cNLL / cECE       \\ \midrule
node-BNNs      & $1.39 \textcolor{gray}{\pm 0.01}$  & $67.6 \textcolor{gray}{\pm 0.3}$           & $0.114 \textcolor{gray}{\pm 0.004}$ & 30.4 / 3.40 / $\mathbf{0.05}$  \\
SWAG           & $1.39 \textcolor{gray}{\pm 0.01}$  & $66.6 \textcolor{gray}{\pm 0.3}$           & $\mathbf{0.020 \textcolor{gray}{\pm 0.005}}$ & 28.4 / 3.72 / 0.11 \\ \midrule
Deep ensembles & $\mathbf{1.15 \textcolor{gray}{\pm 0.00}}$  & $71.6 \textcolor{gray}{\pm 0.0}$           & $0.035 \textcolor{gray}{\pm 0.002}$ & 31.8 / 3.38 / 0.09 \\
weight-RDE     & $\mathbf{1.15 \textcolor{gray}{\pm 0.01}}$  & $71.5 \textcolor{gray}{\pm 0.0}$           & $0.036 \textcolor{gray}{\pm 0.003}$ & 31.7 / 3.39 / 0.09 \\
function-RDE   & $1.21 \textcolor{gray}{\pm 0.02}$  & $70.2 \textcolor{gray}{\pm 0.5}$           & $0.036 \textcolor{gray}{\pm 0.004}$ & 31.1 / 3.43 / 0.10  \\
feature-RDE    & $1.24 \textcolor{gray}{\pm 0.01}$  & $\mathbf{72.0 \textcolor{gray}{\pm 0.1}}$           & $0.100 \textcolor{gray}{\pm 0.003}$ & 31.9 / 3.35 / 0.09 \\ \midrule
LIT & $\mathbf{1.15 \textcolor{gray}{\pm 0.00}}$  & $71.5 \textcolor{gray}{\pm 0.0}$           & $0.035 \textcolor{gray}{\pm 0.002}$ & 31.2 / 3.40 / 0.11 \\
FoRDE-PCA (ours)   & $1.16 \textcolor{gray}{\pm 0.00}$  & $71.4 \textcolor{gray}{\pm 0.0}$           & $0.033 \textcolor{gray}{\pm 0.002}$ & $\mathbf{32.2}$ / $\mathbf{3.28}$ / 0.08 \\ 
\bottomrule
\end{tabular}
\end{sc}
}
\end{table}

\section{Performance under different ensemble sizes}
\begin{figure}[ht]
    \centering
    \begin{subfigure}[b]{0.45\textwidth}
         \centering
         \includegraphics[width=\textwidth]{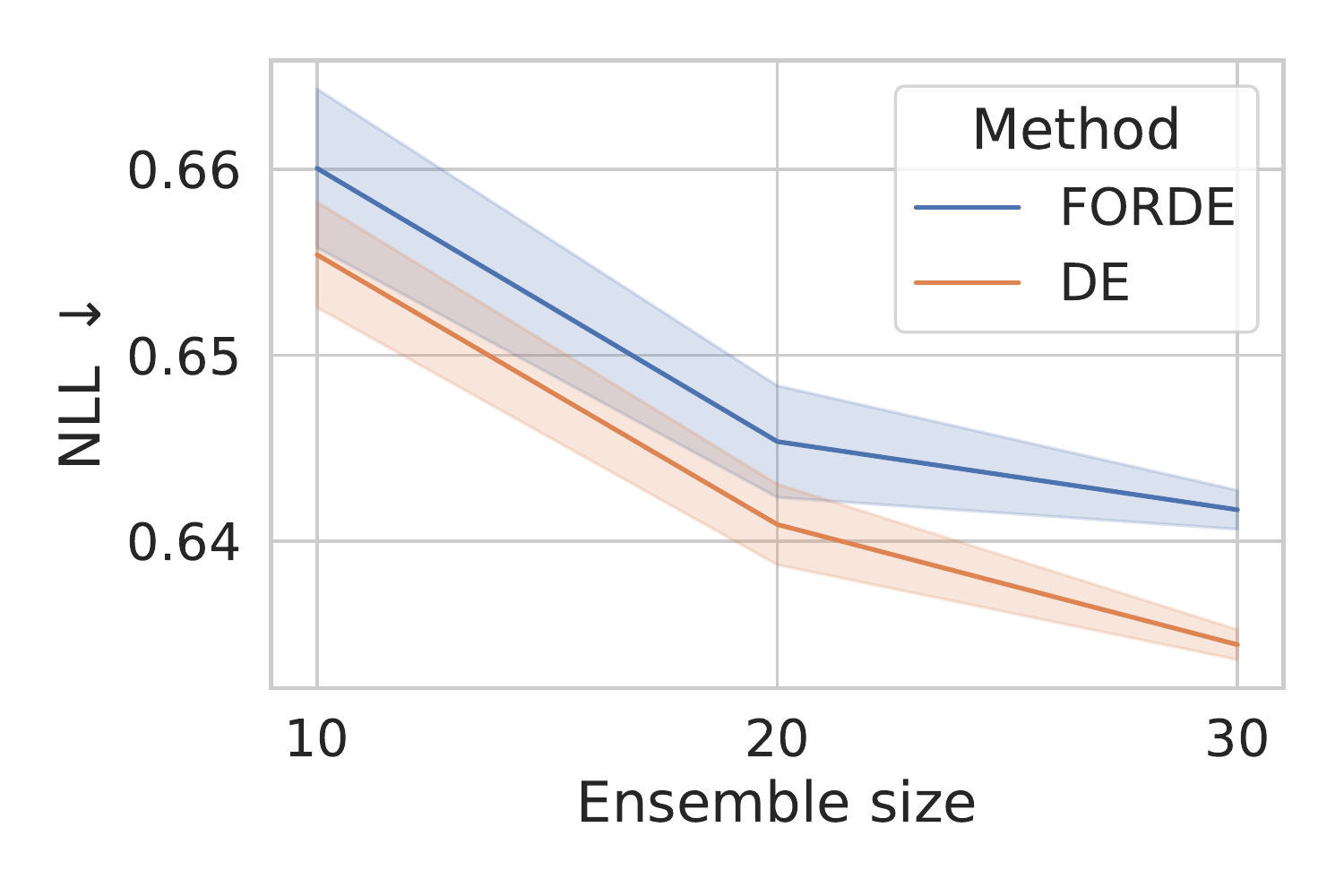}
         \caption{\textsc{cifar-100} (clean)}
         \label{fig:wrn16x4_cifar100_clean_nll_vs_ensemblesize}
    \end{subfigure}
    \hfill
    \begin{subfigure}[b]{0.45\textwidth}
         \centering
         \includegraphics[width=\textwidth]{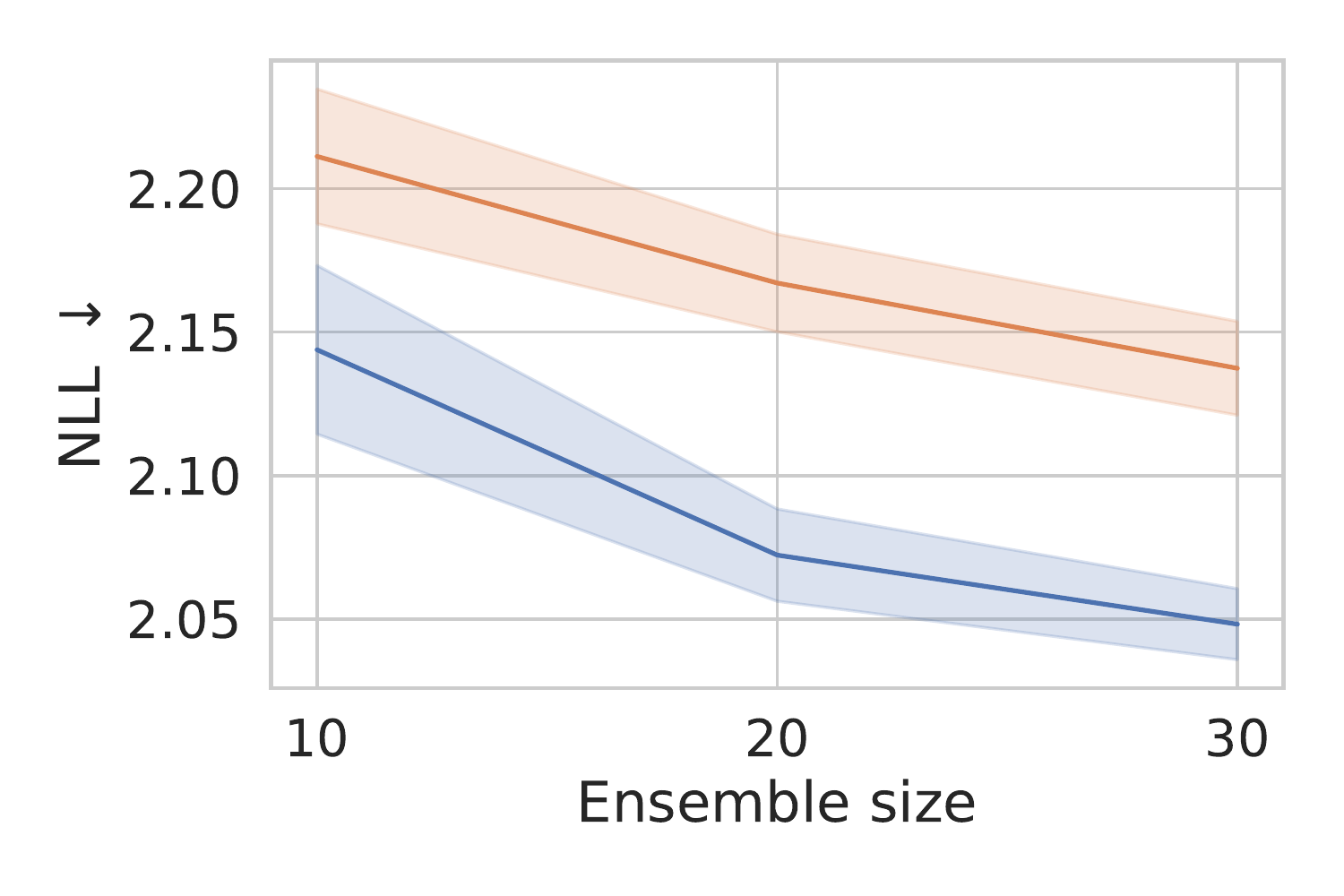}
         \caption{\textsc{cifar-100-C} (corrupted)}
         \label{fig:wrn16x4_cifar100_corrupt_nll_vs_ensemblesize}
    \end{subfigure}
    \caption{\textbf{FoRDE is competitive on in-distribution and outperforms DEs under domain shifts by corruption}. Performance of \textsc{wideresnet16x4} on \textsc{cifar-100} over 5 seeds.}
    \label{fig:wrn16x4_cifar100_nll_vs_ensemblesize}
\end{figure}

\begin{figure}[ht]
    \centering
    \begin{subfigure}[b]{0.45\textwidth}
         \centering
         \includegraphics[width=\textwidth]{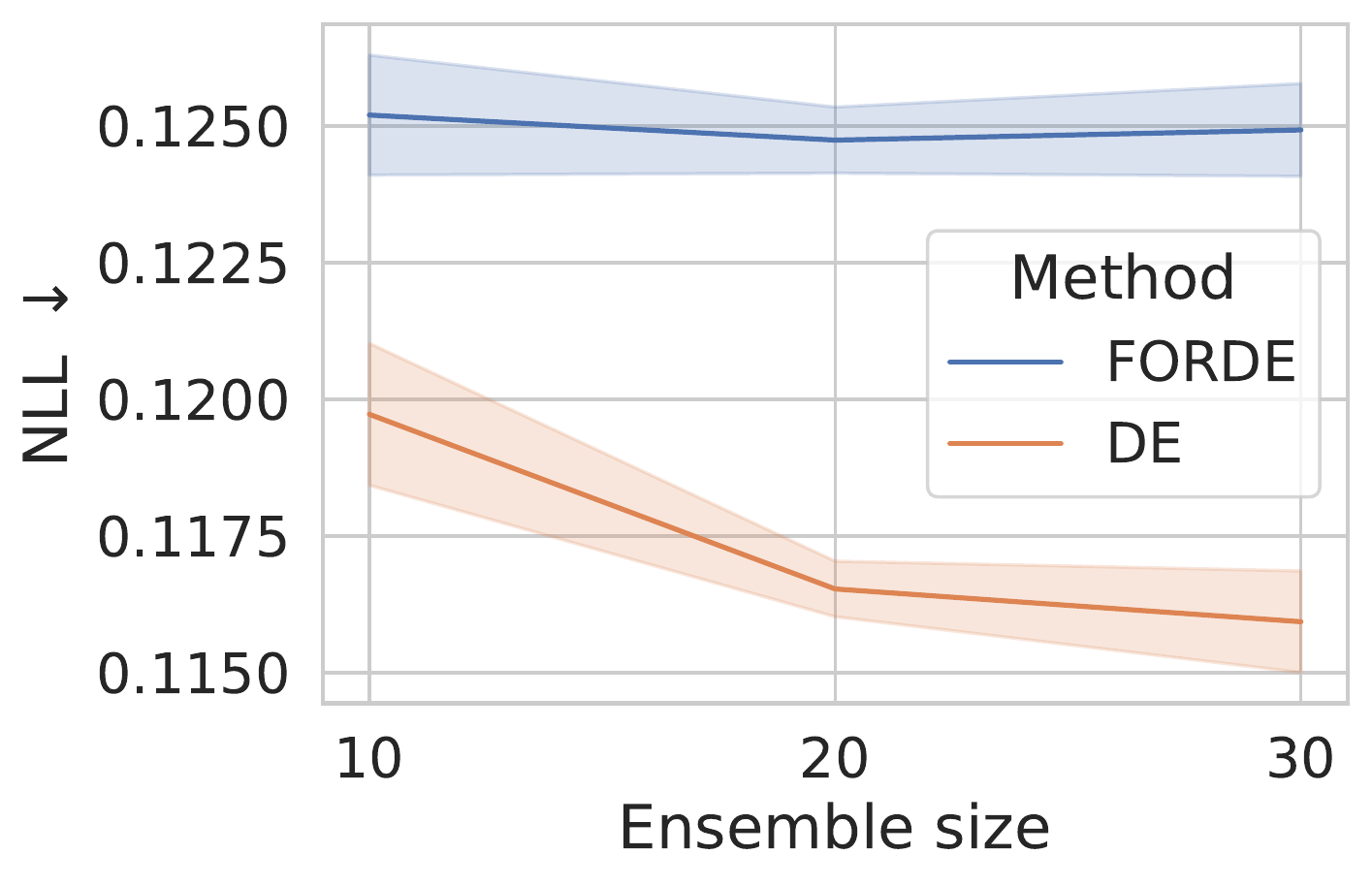}
         \caption{\textsc{cifar-10} (clean)}
         \label{fig:wrn16x4_cifar10_clean_nll_vs_ensemblesize}
    \end{subfigure}
    \hfill
    \begin{subfigure}[b]{0.45\textwidth}
         \centering
         \includegraphics[width=\textwidth]{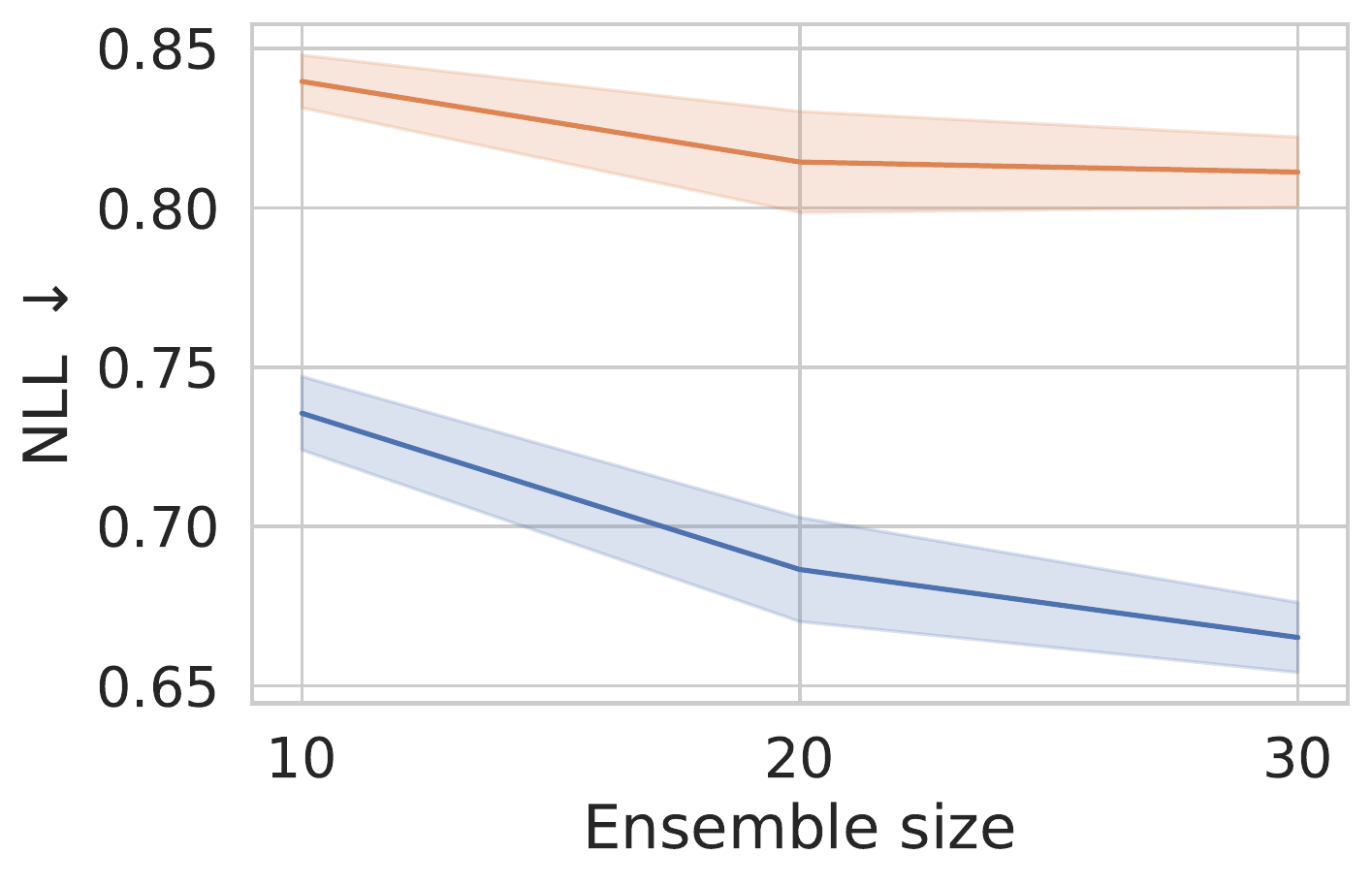}
         \caption{\textsc{cifar-10-C} (corrupted)}
         \label{fig:wrn16x4_cifar10_corrupt_nll_vs_ensemblesize}
    \end{subfigure}
    \caption{\textbf{FoRDE is competitive on in-distribution and outperforms DEs under domain shifts by corruption}. Performance of \textsc{wideresnet16x4} on \textsc{cifar-10} over 5 seeds.}
    \label{fig:wrn16x4_cifar10_nll_vs_ensemblesize}
\end{figure}

We report the NLL of FoRDE and DE under different ensemble sizes on \textsc{cifar-10/100} and \textsc{cifar-10/100-c} in \crefrange{fig:wrn16x4_cifar100_nll_vs_ensemblesize}{fig:wrn16x4_cifar10_nll_vs_ensemblesize}. We use the \textsc{wideresnet16x4} \citep{zagoruyko2016wide} architecture for this experiment. These figures show that both methods enjoy significant improvements in performance as the ensemble size increases. While \cref{fig:wrn16x4_cifar100_clean_nll_vs_ensemblesize} and \cref{fig:wrn16x4_cifar10_clean_nll_vs_ensemblesize} show that FoRDE underperforms DE on clean images, \cref{fig:wrn16x4_cifar100_corrupt_nll_vs_ensemblesize} and \cref{fig:wrn16x4_cifar10_corrupt_nll_vs_ensemblesize} show that FoRDE significantly outperforms DE on corrupted images, such that a FoRDE with 10 members has the same or better corruption robustness of a DE with 30 members.


\section{Training procedure}
\label{algorithm}
\begin{algorithm}[htbp]
   \caption{FoRDE}
   \label{alg:forde}
\begin{algorithmic}[1]
   \State {\bfseries Input:} training data $\mathcal{D}$, orthonormal basis $\mathbf{U}$, diagonal matrix of squared lengthscales $\bS$, a neural network ensemble $\{f(\cdot; \theta_i)\}_{i=1}^M$ of size $M$, positive scalar $\epsilon$, number of iterations $T$, step sizes $\{\eta_t\}_{t=1}^T$, weight decay $\lambda$
   \State {\bfseries Output:} optimized parameters $\{\theta^{(T)}_i\}_{i=1}^M$
   \State Initialize parameters $\{\theta^{(0)}_i\}_{i=1}^M$
   \For{$t=1$ {\bfseries to} $T$}
   \State Draw a mini-batch $\{\x_b, y_b\}_{b=1}^B \sim \mathcal{D}.$
   \For{$b=1$ {\bfseries to} $B$}
        \For{$i=1$ {\bfseries to} $M$} \Comment{Calculate the normalized input gradients for each $\theta_i$ (\cref{eq:base_kernel})}
            \State \begin{equation}
              \mathbf{s}_{i,b} \longleftarrow \frac{\nabla_{\x_b} f\big(\x_b; \theta_i^{(t)}\big)_{y_b}}{\sqrt{\big|\big|\nabla_{\x_b} f\big(\x_b; \theta_i^{(t)}\big)_{y_b}\big|\big|_2^2 + \epsilon^2}}
            \end{equation}
        \EndFor
        \For{$i=1$ {\bfseries to} $M$} \Comment{Calculate the pairwise squared distance in \cref{eq:pca_kernel} }
            \For{$j=1$ {\bfseries to} $M$}
                \begin{equation}
                    d_{i,j,b} \longleftarrow \frac{1}{2} (\mathbf{s}_{i,b}-\mathbf{s}_{j,b})^\top \mathbf{U} \bS \mathbf{U}^\top (\mathbf{s}_{i,b}-\mathbf{s}_{j,b})
                \end{equation}
            \EndFor
        \EndFor
        \State Calculate the global bandwidth per batch sample using the median heuristic (\cref{eq:median_heuristic}):
        \begin{equation}
            h_b \longleftarrow \mathrm{median}\big(\{d_{i,j,b}\}_{i=1,j=1}^{M,M}\big)/(2\ln M)
        \end{equation}
    \EndFor
    \For{$i=1$ {\bfseries to} $M$} \Comment{Calculate the pairwise kernel similarity using \cref{eq:input_grad_minibatch_kernel} and \cref{eq:median_heuristic}}
        \For{$j=1$ {\bfseries to} $M$}
            \begin{equation}
                k_{i,j} \longleftarrow \frac{1}{B}\sum_{b=1}^B \exp\big(-d_{i,j,b}/h_b \big)
            \end{equation}
        \EndFor
    \EndFor
    \For{$i=1$ {\bfseries to} $M$}
        \State Calculate the gradient of the repulsion term using \cref{eq:kde_loggrad}:
        \begin{equation}
            \mathbf{g}^{\mathrm{rep}}_{i} \longleftarrow \frac{\sum_{j=1}^M \nabla_{\theta^{(t)}_i}k_{i,j}}{\sum_{j=1}^M k_{i,j}}
        \end{equation}
        \State Calculate the gradient $\mathbf{g}^{\mathrm{data}}_{i}$ of the cross-entropy loss with respect to $\theta_i$.
        \State Calculate the update vector in \cref{eq:wgd_kde}:
        \begin{equation}
            \v_i^{(t)} \longleftarrow \frac{1}{B}\left(\mathbf{g}^{\mathrm{data}}_{i} - \mathbf{g}^{\mathrm{rep}}_{i}\right) 
        \end{equation}
        \State Update the parameters and apply weight decay:
        \begin{equation}
            \theta^{(t+1)}_i \longleftarrow \theta^{(t)}_i + \eta_t (\v_i^{(t)} - \lambda \theta_i^{(t)})
        \end{equation}
    \EndFor
   \EndFor
\end{algorithmic}
\end{algorithm}

\subsection{Training algorithm for FoRDE}\label{sec:training_algo}
We describe the training algorithm of FoRDE in \cref{alg:forde}.

\subsection{Experimental details for image classification experiments}\label{sec:image_exp_details}
For all the experiments, we used SGD with Nesterov momentum as our optimizer, and we set the momemtum coefficient to $0.9$.
We used a weight decay $\lambda$ of $5 \times 10^{-4}$ and we set the learning rate $\eta$ to $10^{-1}$.
We used a batch size of 128 and we set $\epsilon$ in \cref{alg:forde} to $10^{-12}$. We used 15 bins to calculate ECE during evaluation.

On \textsc{cifar-10} and \textsc{cifar-100}, we use the standard data augmentation procedure, which includes input normalization, random cropping and random horizontal flipping. We ran each experiments for 300 epochs.
We decreased the learning rate $\eta$ linearly from $10^{-1}$ to $10^{-3}$ from epoch 150 to epoch 270.
For evaluation, we used all available types for corruptions and all levels of severity in \textsc{cifar-10/100-c}.

On \textsc{tinyimagenet}, we use the standard data augmentation procedure, which includes input normalization, random cropping and random horizontal flipping.
We ran each experiments for 150 epochs.
We decreased the learning rate $\eta$ linearly from $10^{-1}$ to $10^{-3}$ from epoch 75 to epoch 135.
For evaluation, we used all available types for corruptions and all levels of severity in \textsc{tinyimagenet-c}.

For weight-RDE and FoRDE, we only imposed a prior on the weights via the weight decay parameter.
For feature-RDE and function-RDE, we followed the recommended priors in \citet{yashima22a}. 
For feature-RDE, we used Cauchy prior with a prior scale of $10^{-3}$ for \textsc{cifar-10} and a prior scale of $5 \times 10^{-3}$ for both \textsc{cifar-100} and \textsc{tinyimagenet}, and we used a projection dimension of $5$.
For function-RDE, we used Cauchy prior with a prior scale of $10^{-6}$ for all datasets.

\subsection{Experimental details for transfer learning experiments}\label{sec:transferlearning_details}
We extracted the outputs of the last hidden layer of a Vision Transformer model pretrained on the ImageNet-21k dataset (\texttt{google/vit-base-patch16-224-in21k} checkpoint in the \texttt{transformers} package from \texttt{huggingface}) and use them as input features, and we trained ensembles of 10 ReLU networks with 3 hidden layers and batch normalization.

For all the experiments, we used SGD with Nesterov momentum as our optimizer, and we set the momemtum coefficient to $0.9$. We used a batch size of 256, and we annealed the learning rate from $0.2$ to $0.002$ during training. We used a weight decay of $5 \times 10^{-4}$. We used 15 bins to calculate ECE for evaluation. For OOD experiments, we calculated epistemic uncertainty on the test sets of \textsc{cifar-10/100} and \textsc{cinic10}. For evaluation on natural corruptions, we used all available types for corruptions and all levels of severity in \textsc{cifar-10/100-c}.

\section{Additional results}

\subsection{Input gradient diversity and functional diversity}
\begin{figure}[ht]
    \centering
    \includegraphics[width=.8\textwidth]{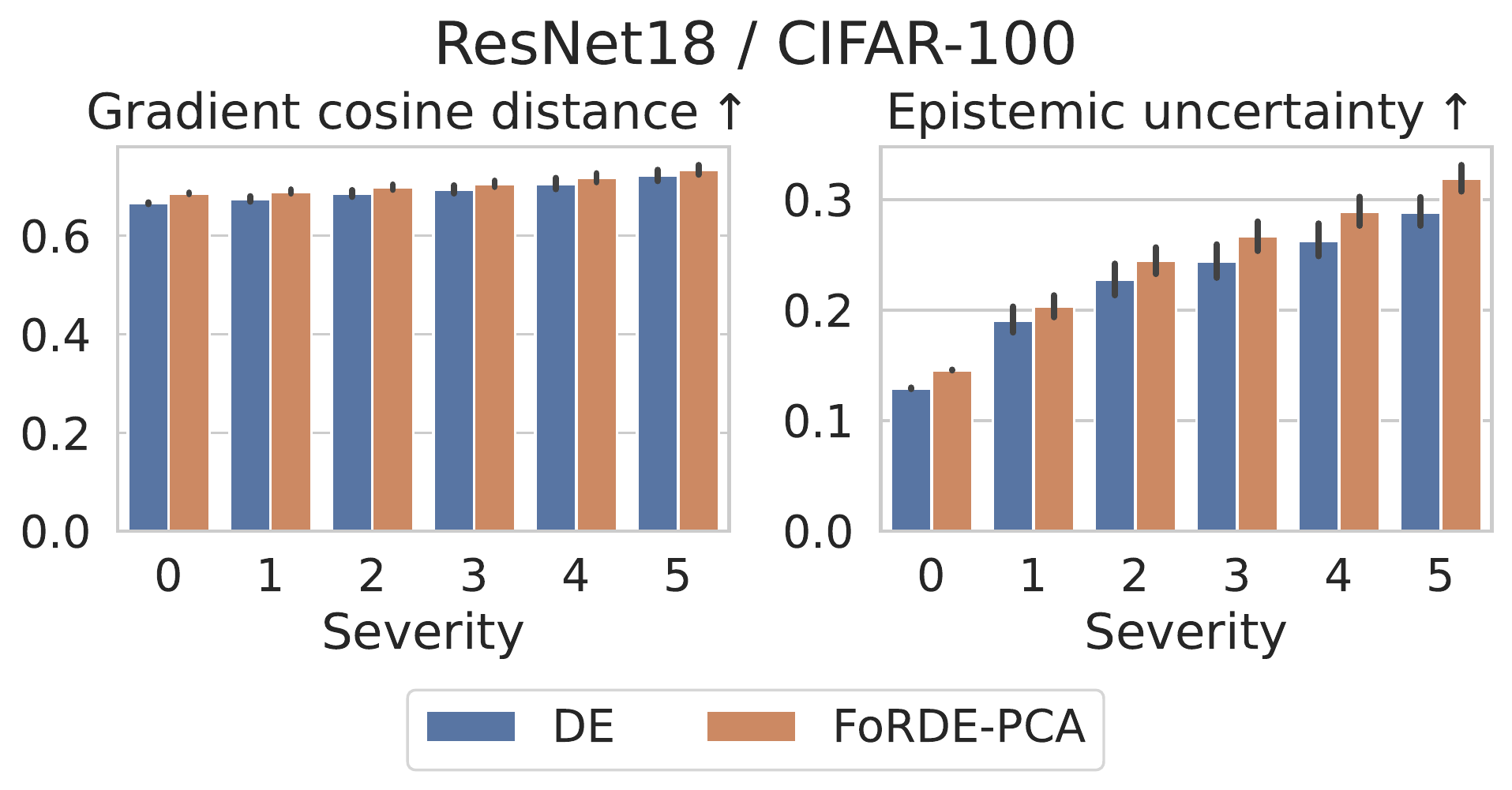}
    \caption{\textbf{FoRDE has higher gradient distance as well as higher epistemic uncertainty} Results of \textsc{resnet18} on \textsc{cifar100} over 5 seeds under different levels of corruption severity, where $0$ mean no corruption.}
    \label{fig:resnet18_cifar100_graddist_epistemic}
\end{figure}

To show that FoRDE indeed produces ensembles with higher input gradient diversity among member models, which in turn leads to higher functional diversity than DE, we visualize the input gradient distance and epistemic uncertainty of FoRDE and DE in \cref{fig:resnet18_cifar100_graddist_epistemic}. 
To measure the differences between input gradients, we use \emph{cosine distance}, defined as $1-\cos(\mathbf{u}, \mathbf{v})$ where $\cos(\mathbf{u}, \mathbf{v})$ is the cosine similarity between two vectors $\mathbf{u}$ and $\mathbf{v}$.
To quantify functional diversity, we calculate the epistemic uncertainty using the formula in \cite{pmlr-v80-depeweg18a}, similar to the transfer learning experiments. 
\cref{fig:resnet18_cifar100_graddist_epistemic} shows that FoRDE has higher gradient distances among members compared to DE, while also having higher epistemic uncertainty across all levels of corruption severity. Intuitively, as the test inputs become more corrupted, epistemic uncertainty of both FoRDE and DE increases, and the input gradients between member models become more dissimilar for both methods. These results suggest that there could be a connection between input gradient diversity and functional diversity in neural network ensembles.

\subsection{Performance under corruptions}
\begin{figure}[ht]
    \centering
    \includegraphics[width=\textwidth]{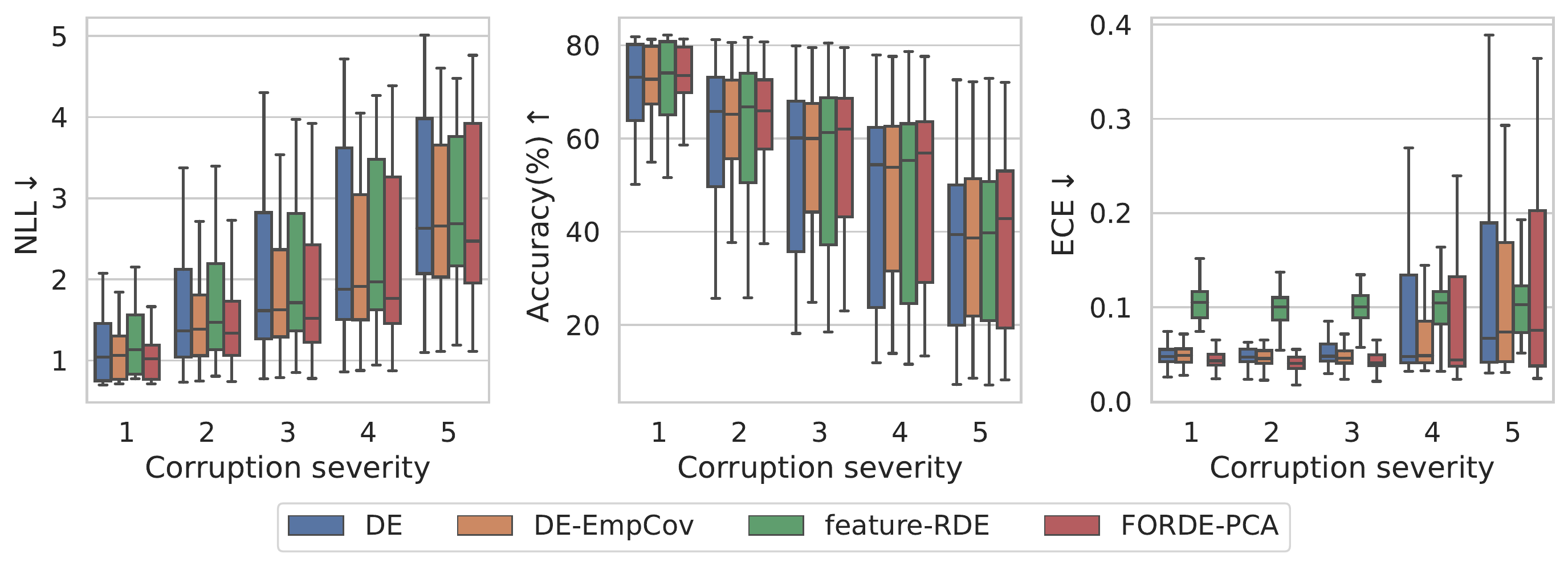}
    \caption{\textbf{FoRDE performs better than baselines across all metrics and under all corruption severities.} Results for \textsc{resnet18}/\textsc{cifar-100-c}. Each ensemble has 10 members.}
    \label{fig:ensemble_corruption_cifar100}
\end{figure}

\begin{figure}[ht]
    \centering
    \includegraphics[width=\textwidth]{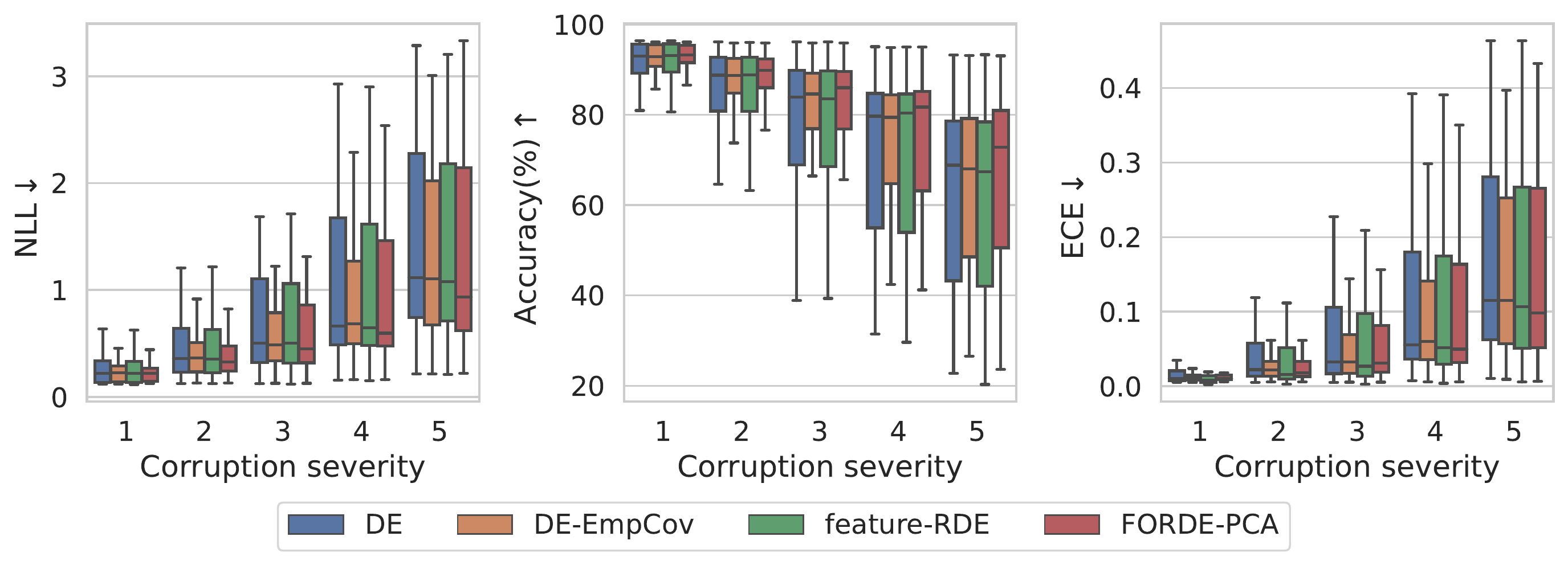}
    \caption{\textbf{FoRDE performs better than baselines across all metrics and under all corruption severities.} Results for \textsc{resnet18}/\textsc{cifar-10-c}. Each ensemble has 10 members.}
    \label{fig:ensemble_corruption_cifar10}
\end{figure}

We plot performance of all methods under the \textsc{resnet18}/\textsc{cifar-c} setting in \cref{fig:ensemble_corruption_cifar100,fig:ensemble_corruption_cifar10}. These figures show that FoRDE achieves the best performance across all metrics under all corruption severities.


\subsection{Comparison between FoRDE-PCA and EmpCov prior}
\begin{figure}[ht]
    \centering
    \includegraphics[width=\textwidth]{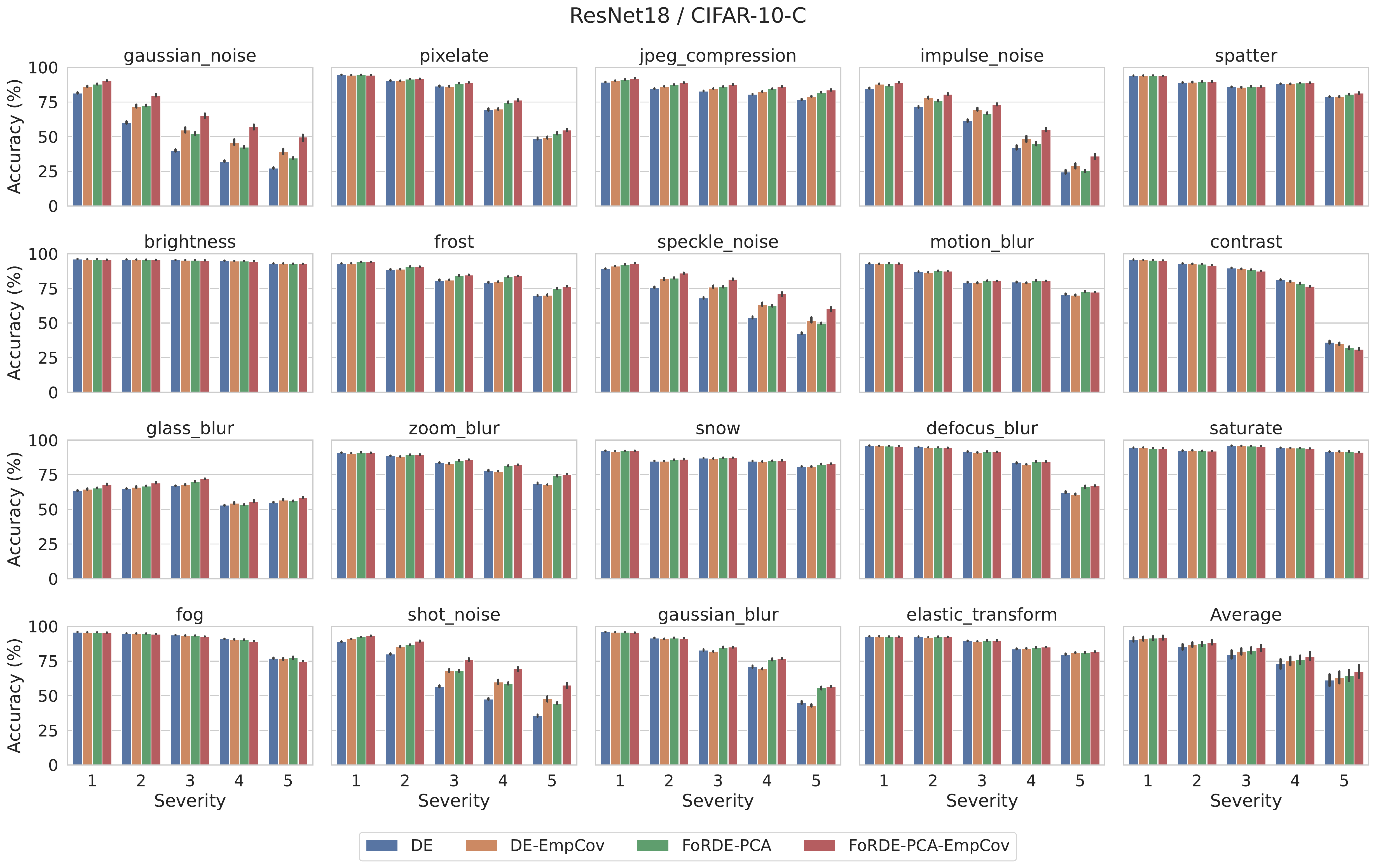}
    \caption{\textbf{FoRDE-PCA and EmpCov prior behave similarly in most of the corruption types} Here we visualize accuracy for each of the 19 corruption types in \textsc{cifar-10-c} in the first 19 panels, while the last panel (bottom right) shows the average accuracy. Both FoRDE-PCA and DE-EmpCov are more robust than plain DE on most of the corruption types, with the exception of \emph{contrast} where both FoRDE-PCA and DE-EmpCov are less robust than DE. On the other hand, on the \emph{blur} corruption types (\emph{\{motion, glass, zoom, defocus, gaussian\}}-blur), DE-EmpCov is less robust than vanilla DE while FoRDE-PCA exhibits better robustness than DE.}
    \label{fig:resnet18_cifar10c_empcov_forde}
\end{figure}

In \cref{sec:pca_lengthscales}, we discussed a possible connection between FoRDE-PCA and the EmpCov prior \citep{izmailov2021dangers}.
Here, we further visualize performance of FoRDE-PCA, DE with EmpCov prior and vanilla DE on different types of corruptions and levels of severity for the \textsc{resnet18}/\textsc{cifar10} setting in \cref{fig:resnet18_cifar10c_empcov_forde}.
This figure also includes results of FoRDE-PCA with EmpCov prior to show that these two approaches can be combined together to further boost corruption robustness of an ensemble.
Overall, \cref{fig:resnet18_cifar10c_empcov_forde} shows that FoRDE-PCA and DE-EmpCov have similar behaviors on the majority of the corruption types, meaning that if DE-EmpCov is more or less robust than DE on a corruption type then so does FoRDE-PCA.
The exceptions are the \emph{blur} corruption types (\emph{\{motion, glass, zoom, defocus, gaussian\}}-blur), where DE-EmpCov is less robust than vanilla DE while FoRDE-PCA exhibits better robustness than DE.
Finally, by combining FoRDE-PCA and EmpCov prior together, we achieve the best robustness on average.

\subsection{Tuning the lengthscales for the RBF kernel}\label{sec:tuning_lengthscales}
\begin{figure}[ht]
    \centering
    \includegraphics[width=\textwidth]{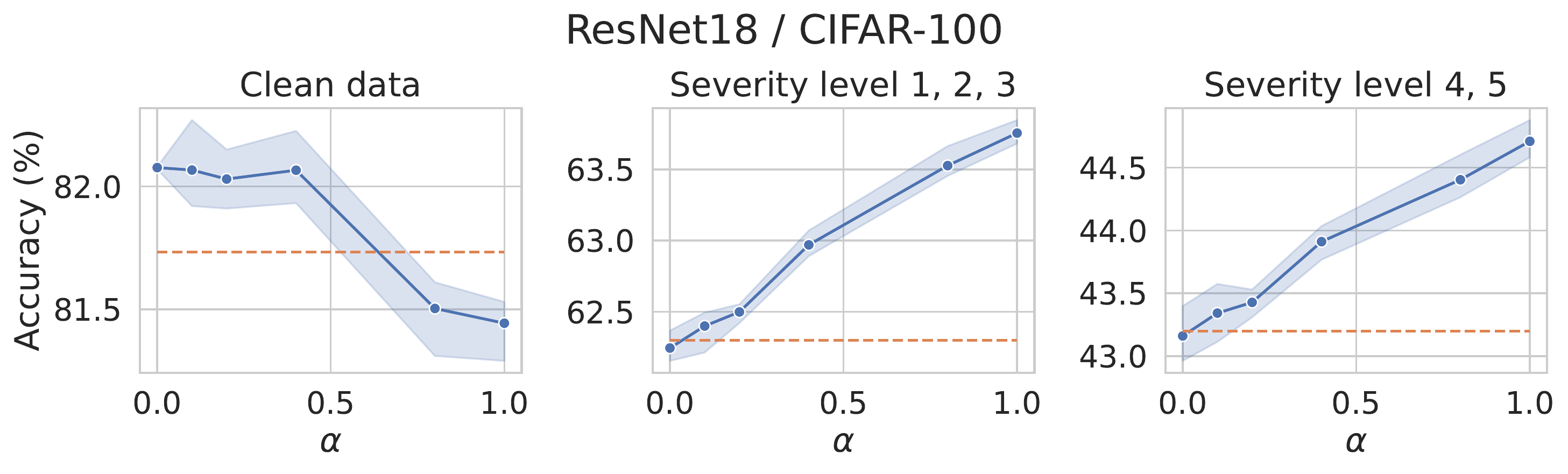}
    \caption{\textbf{When moving from the identity lengthscales to the PCA lengthscales, FoRDE becomes more robust against natural corruptions, while exhibiting small performance degradation on clean data.} Results are averaged over 3 seeds. Blue lines show performance of FoRDE, while orange dotted lines indicate the average accuracy of DE for comparison. At the identity lengthscales, FoRDE has higher accuracy than DE on in-distribution data but are slightly less robust against corruptions than DE. As we move from the identity lengthscales to the PCA lengthscales, FoRDE becomes more and more robust against corruptions, while showing a small decrease in in-distribution performance. Here we can see that $\alpha=0.4$ achieves good balance between in-distribution accuracy and corruption robustness.}
    \label{fig:resnet18_cifar100_alpha_lengthscale}
\end{figure}

\begin{figure}[ht]
    \centering
    \includegraphics[width=\textwidth]{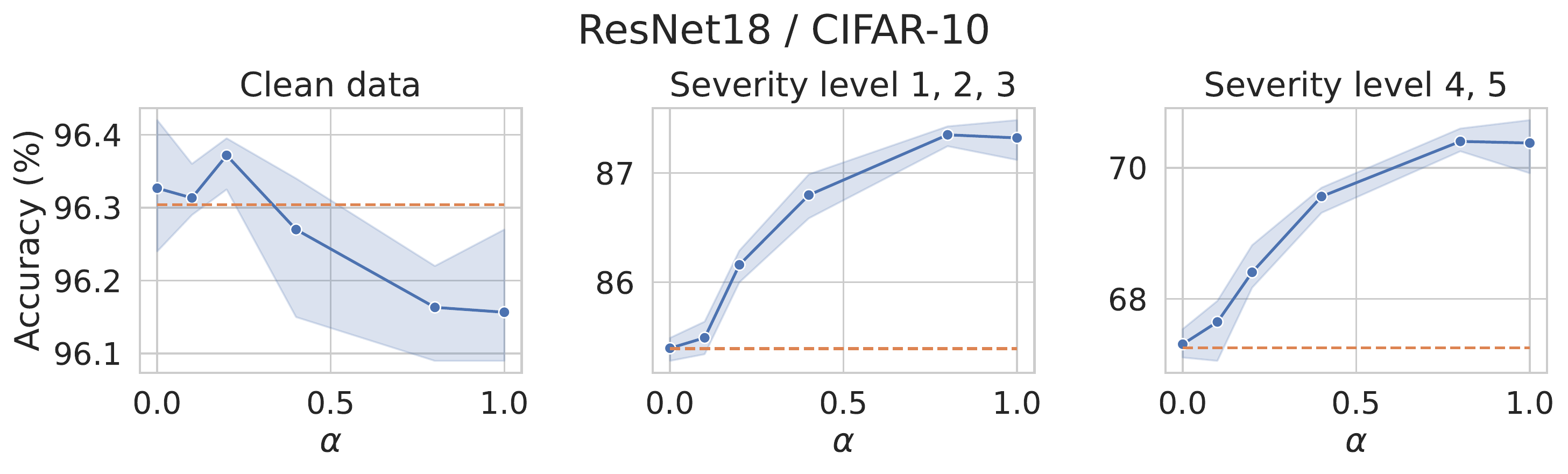}
    \caption{\textbf{When moving from the identity lengthscales to the PCA lengthscales, FoRDE becomes more robust against natural corruptions, while exhibiting small performance degradation on clean data.} Results are averaged over 3 seeds. Blue lines show performance of FoRDE, while orange dotted lines indicate the average accuracy of DE for comparison. At the identity lengthscales, FoRDE has higher accuracy than DE on in-distribution data but are slightly less robust against corruptions than DE. As we move from the identity lengthscales to the PCA lengthscales, FoRDE becomes more and more robust against corruptions, while showing a small decrease in in-distribution performance. Here we can see that $\alpha=0.2$ achieves good balance between in-distribution accuracy and corruption robustness.}
    \label{fig:resnet18_cifar10_alpha_lengthscale}
\end{figure}

In this section, we show how to tune the lengthscales for the RBF kernel by taking the weighted average of the identity lengthscales and the PCA lengthscales introduced in \cref{sec:pca_lengthscales}.
Particularly, using the notation of \cref{sec:pca_lengthscales}, we define the diagonal lengthscale matrix $\bS_\alpha$:
\begin{equation}
    \bS_\alpha = \alpha \boldsymbol{\Lambda}^{-1} + (1-\alpha)\mathbf{I}
\end{equation}
where $\boldsymbol{\Lambda}$ is a diagonal matrix containing the eigenvalues from applying PCA on the training data as defined in \cref{sec:pca_lengthscales}.
We then visualize the accuracy of FoRDE-PCA trained under different $\alpha \in \{0.0, 0.1, 0.2, 0.4, 0.8, 1.0\}$ in \cref{fig:resnet18_cifar100_alpha_lengthscale} for the \textsc{resnet18}/\textsc{cifar-100} setting and in \cref{fig:resnet18_cifar10_alpha_lengthscale} for the \textsc{resnet18}/\textsc{cifar-10} setting.
\cref{fig:resnet18_cifar100_alpha_lengthscale} shows that indeed we can achieve good performance on both clean and corrupted data by choosing a lengthscale setting somewhere between the identity lengthscales and the PCA lengthscales, which is at $\alpha=0.4$ in this experiment. A similar phenomenon is observed in \cref{fig:resnet18_cifar10_alpha_lengthscale}, where $\alpha=0.2$ achieves the best results on both clean and corrupted data.

\end{document}